\RequirePackage[svgnames]{xcolor}

\documentclass[11pt,letterpaper]{mystyle}

\usepackage[all]{hypcap}
\usepackage[svgnames]{xcolor}
\usepackage[comma,authoryear,compress]{natbib}
\usepackage{algorithm}
\usepackage{algorithmicx}
\usepackage{algpseudocode}
\usepackage{microtype}
\usepackage{graphicx}
\expandafter\def\csname ver@subfig.sty\endcsname{}
\usepackage{booktabs} %
\usepackage{float}
\usepackage{bigstrut}

\usepackage{amsmath}
\usepackage{amssymb}
\usepackage{mathtools}
\usepackage{amsthm}
\usepackage{mathrsfs}
\usepackage{nicefrac}
\usepackage{dsfont}
\usepackage{enumitem}
\usepackage{subcaption}
\usepackage{graphicx,subfig}
\usepackage{cleveref}
\usepackage{bxcoloremoji}

\usepackage{float}

\usepackage[utf8]{inputenc} %
\usepackage[T1]{fontenc}    %
\usepackage{url}            %
\usepackage{booktabs}       %
\usepackage{amsfonts}       %
\usepackage{nicefrac}       %
\usepackage{microtype}      %
\usepackage{graphicx}
\usepackage{subcaption}
\usepackage{amssymb}
\usepackage{fdsymbol}
\usepackage{wrapfig}
\usepackage{lipsum}
\usepackage{enumitem}
\usepackage{stackengine}
\usepackage[font=small,labelfont=bf]{caption}
\usepackage{color}
\usepackage{adjustbox}

\usepackage{rotating}
\usepackage{makecell}

\newcommand{\x}{\mathbf{x}}

\DeclareMathOperator*{\argmin}{argmin}

\definecolor{blanchedalmond}{rgb}{1.0, 0.92, 0.8}
\definecolor{carmine}{rgb}{0.59, 0.0, 0.09}
\definecolor{lightblue}{rgb}{0.22,0.45,0.70}%

\renewcommand{\mathbf}{\boldsymbol}

\makeatletter
\def\Ddots{\mathinner{\mkern1mu\raise\p@
\vbox{\kern7\p@\hbox{.}}\mkern2mu
\raise4\p@\hbox{.}\mkern2mu\raise7\p@\hbox{.}\mkern1mu}}
\makeatother

\definecolor{amaranth}{rgb}{0.9, 0.17, 0.31}
\definecolor{antiquebrass}{rgb}{0.8, 0.58, 0.46}
\definecolor{antiquefuchsia}{rgb}{0.57, 0.36, 0.51}
\definecolor{chromeyellow}{rgb}{0.31, 0.47, 0.26}

\newcommand{\github}{\raisebox{-1.5pt}{\includegraphics[height=1.05em]{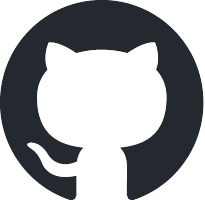}}}
\newcommand{\wnb}{\raisebox{-1.5pt}{\includegraphics[height=1.05em]{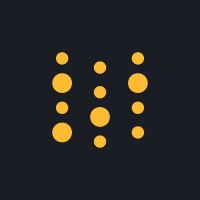}}}
\newcommand{\resa}{\raisebox{-1.5pt}{\includegraphics[height=1.05em]{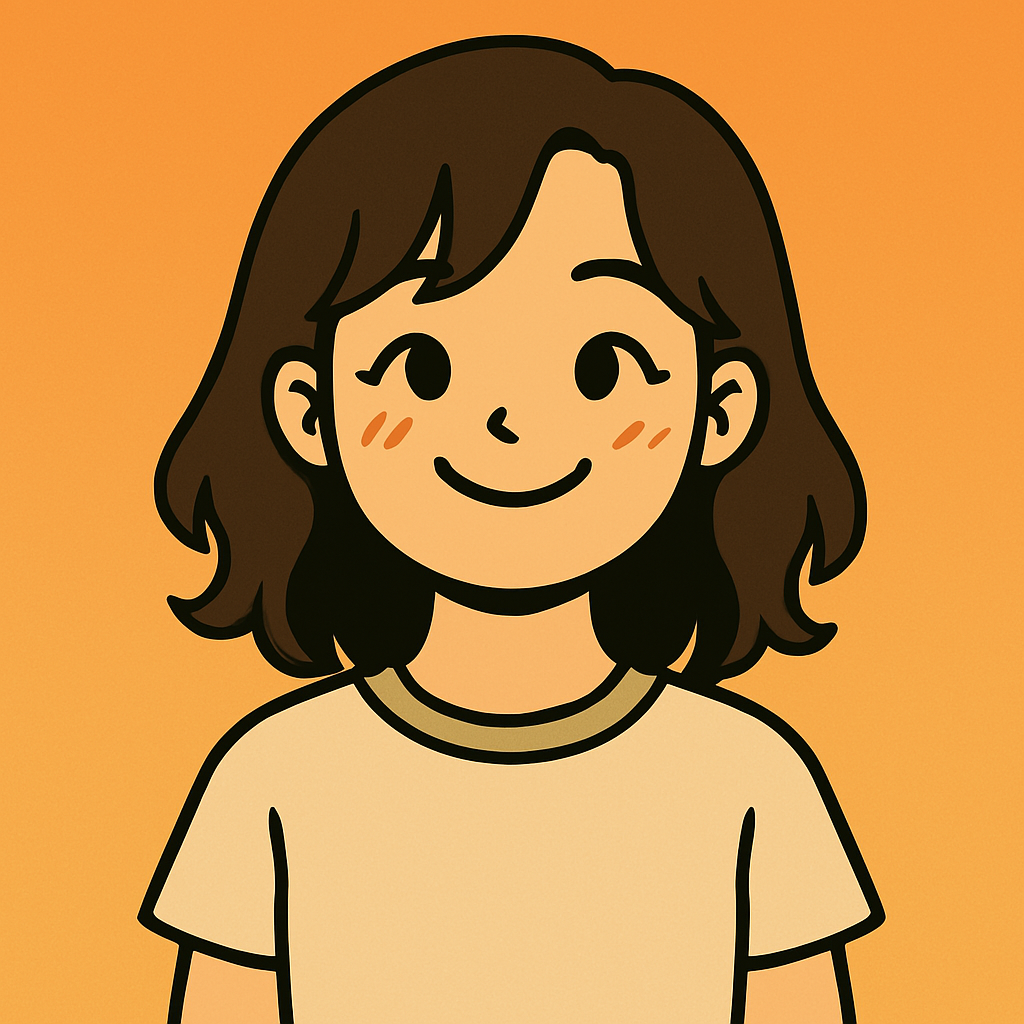}}}

\newtcolorbox{AIbox}[2][]{aibox,title=#2,#1}
\definecolor{lightblue}{rgb}{0.22,0.45,0.70}%
\definecolor{Gray}{gray}{0.95}
\definecolor{Cornsilk}{rgb}{1.0, 0.97, 0.86}

\usepackage{amsmath}

\usepackage[all]{hypcap}

\title{{\textbf{Resa}}: Transparent Reasoning Models via SAEs}
\runningtitle{Resa: Transparent Reasoning Models via SAEs}

\author[ ]{% Ensure no unwanted space at the beginning
  Shangshang Wang,  % Add affiliation marker and line break
  Julian Asilis, 
  Ömer Faruk Akgül,
  Enes Burak Bilgin, \\
  Ollie Liu, 
  Deqing Fu, and 
  Willie Neiswanger
}

\affil[ ]{University of Southern California}

\correspondingauthor{Shangshang Wang \href{https://shangshang-wang.github.io}{shangshangwang.github.io}; Willie Neiswanger \href{mailto:neiswang@usc.edu}{neiswang@usc.edu}}

\begin{document}

\begin{abstract}

How cost-effectively can we elicit strong reasoning in language models by leveraging their underlying representations? We answer this question with Resa, a family of 1.5B reasoning models trained via a novel and efficient sparse autoencoder tuning (SAE‑Tuning) procedure. This method first trains an SAE to capture reasoning abilities from a source model, and then uses the trained SAE to guide a standard supervised fine-tuning process to elicit such abilities in a target model, all using verified question-answer data \textit{without any reasoning traces}. Notably, when applied to certain base models before further RL post-training, SAE‑Tuning retains >97\% of its RL‑trained counterpart’s reasoning performance while reducing training costs by >2000x to roughly \$1 and training time by >450x to around 20 minutes. Furthermore, when applied to lightly RL‑trained models (e.g., within 1 hour on 2 GPUs), it enables reasoning performance such as 43.33\% Pass@1 on AIME24 and 90\% Pass@1 on AMC23 for only around \$1 additional cost. Surprisingly, the reasoning abilities extracted via SAEs are potentially both generalizable and modular. Generality means abilities extracted from one dataset still elevate performance on a larger and overlapping corpus. Modularity means abilities extracted from Qwen or Qwen‑Math can be attached to the R1‑Distill model at test time, \textit{without any retraining}, and yield comparable gains. Extensive ablations validate these findings and all artifacts are fully open-sourced.

%%%% OLDER VERSION (5am)
% How cost-effectively can one elicit strong reasoning abilities if one knows where and how to extract them? We answer this question with Resa, a family of 1.5B reasoning models trained via a novel and transparent sparse autoencoder tuning (SAE‑Tuning) procedure. It first trains an SAE to extract reasoning abilities from a source model using verified question-answer data \textit{without any} reasoning traces, and then uses the trained SAE to guide a standard supervised fine-tuning process to elicit such abilities in a target model using the same data \textit{without any} curation. Notably, when applied to certain base models before further RL post-training, SAE‑Tuning retains >97\% of its RL‑trained counterpart’s reasoning performance while reducing training costs by >2000x to roughly \$1 and training time by >450x to around 20 minutes. Furthermore, when applied to lightly RL‑trained models (e.g., within 1 hour on 2 GPUs), it enables reasoning performance such as 43.33\% Pass@1 on AIME24 and 90\% Pass@1 on AMC23 for only around \$1 additional cost. Surprisingly, the reasoning abilities extracted via SAEs are potentially both generalizable and modular. Generality means abilities extracted from one dataset still elevate performance on a larger and overlapping corpus. Modularity means abilities extracted from Qwen or Qwen‑Math can be attached to the R1‑Distill model at test time, \textit{without any} retraining, and yield comparable gains. Extensive ablations validate these findings and all artifacts are fully open-sourced.

\vspace{5mm}

\resa{} \textbf{Notion Blog}: \href{https://shangshangwang.notion.site/resa}{https://shangshangwang.notion.site/resa}

\github{} \textbf{Code Repository}: \href{https://github.com/shangshang-wang/Resa}{https://github.com/shangshang-wang/Resa}

\wnb{} \textbf{Training Logs}: \href{https://wandb.ai/upup-ashton-wang-usc/Resa}{https://wandb.ai/upup-ashton-wang-usc/Resa}

\coloremojicode{1F917} \textbf{Model Weights \& Checkpoints}: \href{https://huggingface.co/Resa-Yi}{https://huggingface.co/Resa-Yi}
\end{abstract}

\maketitle
\vspace{3mm}
\vspace{-4mm}
\section{Introduction}
\label{sec:intro}

Reasoning language models have demonstrated increasing performance in domains like math, coding, and science~\citep{wangllmreasoning2025, xu2025towards}. Despite the impressive reasoning performance elicited by reinforcement learning (RL) or supervised fine-tuning (SFT)~\citep{chu2025sftmemorizesrlgeneralizes}, these methods often operate as a ``black box''. In other words, while they improve reasoning, how they alter the model's internal representations to do so is largely opaque. Furthermore, RL-based workflows are notoriously resource-intensive, requiring vast computational power and long training time to converge. On the other hand, SFT hinges on the availability of high-quality  Chain-of-Thought (CoT) reasoning traces, which are costly to curate~\citep{muennighoff2025s1simpletesttimescaling}. This leaves a critical gap in the field: The need for a three-birds-one-stone method that can elicit strong reasoning abilities in a way that is not only \textit{effective} but also computationally \textit{efficient} and \textit{transparent}.

\begin{figure}[!t]
    \centering
    \includegraphics[width=.95\linewidth]{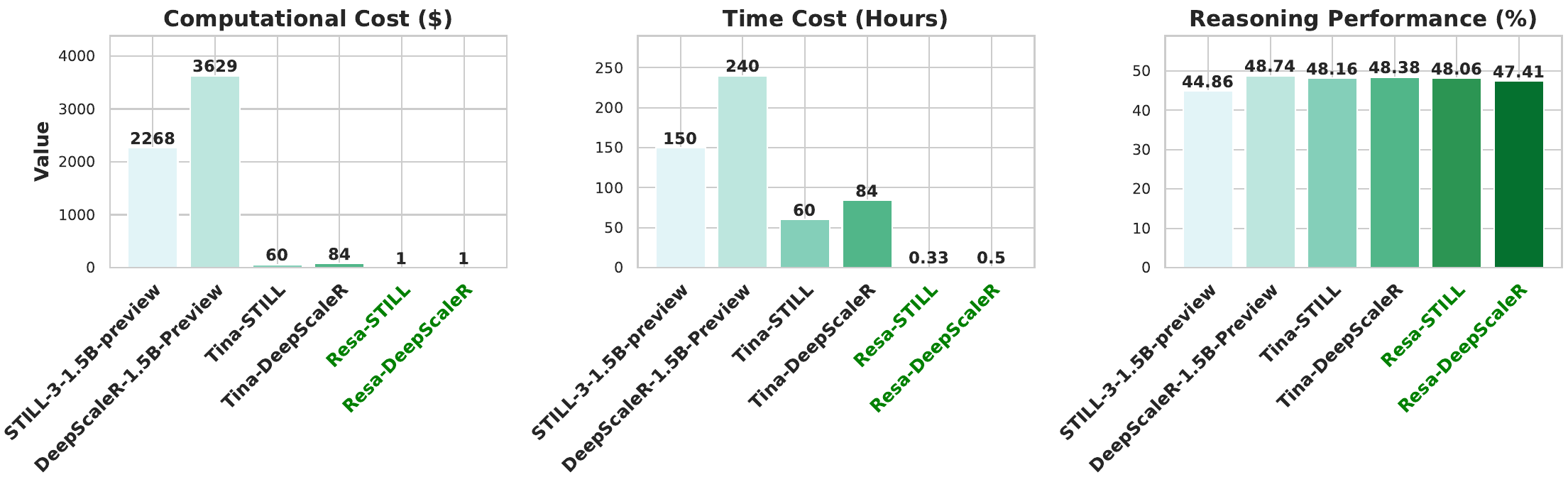}
    \caption{\textbf{Comparison between Example Resa Models and Baselines}\; The Tina models correspond to the best checkpoints in~\citet{wang2025tinatinyreasoningmodels}. Resa-STILL and Resa-DeepScaleR correspond to Resa-STILL-v5 and Resa-DeepScaleR-v3 in Table~\ref{tab:reasoning elicitation}, respectively. For these Resa models, the required SAEs are trained from scratch (as shown in Section~\ref{sec:practical utility}) and both computational and time costs are total costs for training SAEs and models. Reasoning performance denotes the average zero-shot Pass@1 score across AIME24/25, AMC23, MATH500, GPQA Diamond, and Minerva benchmarks.}
    \label{fig:overall_comparison}
\end{figure}

In this paper, we bridge this gap with \textit{Resa}, a family of 1.5B \underline{re}asoning models trained via sparse autoencoders (\underline{SA}Es), using a novel \textit{SAE-Tuning} procedure. SAEs are unsupervised models designed to deconstruct a model's dense internal activations into a sparse dictionary of more interpretable latent features~\citep{anthropic2023features, anthropic2024scaling}. Our key insight is that within this dictionary, certain features must correspond to the fundamental building blocks of reasoning. By instilling latent reasoning features captured by an SAE back into a model via a tuning procedure, we can effectively and efficiently elicit the model's reasoning abilities.

Specifically, SAE-Tuning involves two key stages: First, we use an SAE to probe the internal activations of a source model, identifying and extracting a dictionary of latent features that correspond to its reasoning processes. Second, we freeze this feature-rich SAE and insert it into a target model to guide a SFT process to elicit reasoning abilities in the target model. SAE-Tuning also distinguishes itself from existing methods by using SFT on a minimal verified CoT-free question-answer data type. By verified, we mean that the answer correctness is ensured via methods like human annotation or language-model-based verification~\citep{guha2025openthoughtsdatarecipesreasoning}, while CoT-free signifies that our SAE-Tuning procedure functions without needing explicit step-by-step reasoning traces. Crucially, our control experiments in Table~\ref{tab:reasoning elicitation} demonstrate that performing standard SFT on this same CoT-free data without an SAE fails to elicit any meaningful reasoning, highlighting the vital role of the SAE. We summarize our core contributions as follows:
\begin{itemize}[leftmargin=7mm,itemsep=3mm, topsep=2mm]
    \item \textbf{Efficient Reasoning Ability Elicitation}\; Purely using verified CoT-free data, we demonstrate that SAE-Tuning can be applied in an end-to-end manner to certain base models with a trained-from-scratch SAE to elicit reasoning abilities on par with those achieved via costly RL. This leads to substantial gains with peak training cost reductions of over 2000x (to approximately \$1) and time reductions of over 450x (to under 20 minutes) compared to RL-based workflows, while maintaining comparable performance.
    \item \textbf{Generalizable and Modular Reasoning Ability}\; We establish the generality and modularity of the extracted reasoning abilities such that these abilities generalize across out-of-distribution datasets and can be attached to models within the same family at test time without additional training, functioning as a portable reasoning adapter.
    \item \textbf{Transparent Reasoning Feature Extraction}\; We provide a transparent view into the model's reasoning abilities. Specifically, we propose a novel prompt-only method to extract and quantify latent reasoning features using SAEs and show that the layer-wise distribution of these reasoning features correlates with reasoning performance of Resa models, offering a data-driven path to optimizing SAE-Tuning.
\end{itemize}

\section{Resa: Transparent Reasoning Models via SAEs}
\label{sec:resa}

Resa is a family of 1.5B transparent reasoning models derived from the Tina models~\citep{wang2025tinatinyreasoningmodels} and R1-Distill\footnote{\href{https://huggingface.co/deepseek-ai/DeepSeek-R1-Distill-Qwen-1.5B}{deepseek-ai/DeepSeek-R1-Distill-Qwen-1.5B}}~\citep{deepseekai2025deepseekr1incentivizingreasoningcapability} via SAEs. The transparent designation reflects our methodology's focus on interpretability. We use an SAE to explicitly isolate and extract implicit reasoning abilities (i.e., latent reasoning features) from a source model, and use that trained SAE to controllably instill those features into a target model to elicit reasoning abilities. The entire procedure relies on a novel SAE-Tuning procedure, which uses only verified CoT-free question-answer data.

\begin{figure}[!h]
    \centering
    \includegraphics[width=.9\linewidth]{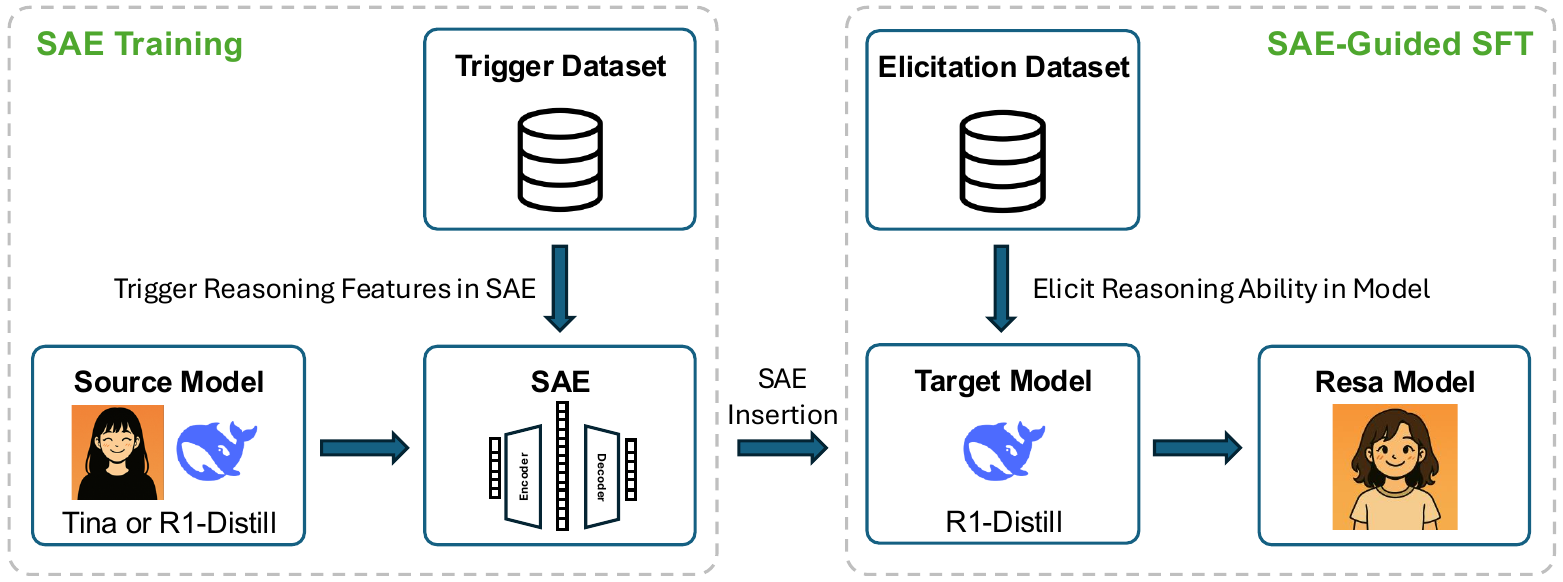}
    \caption{\textbf{Two-Stage Pipeline of SAE-Tuning}\; The procedure begins with SAE training (Left), where an SAE is trained to capture reasoning features from a source model with a trigger dataset. During SAE-guided SFT (Right), the trained SAE is then frozen and inserted into a target model. An elicitation dataset is used to guide a SFT process to elicit the reasoning abilities in the target model. Notably, the trigger and elicitation datasets are usually the same CoT-free data. See Section~\ref{sec:method} for a description of each component.}
    \label{fig:resa pipeline}
\end{figure}

\subsection{Sparse Autoencoder Tuning}
\label{sec:method}

SAE-Tuning is an efficient two-stage training procedure designed to transfer implicit reasoning abilities from a source model to a target model; this full procedure is summarized in Figure~\ref{fig:resa pipeline}. The two stages consist of:
\begin{itemize}[leftmargin=7mm,itemsep=2mm, topsep=0em]
    \item \textbf{Stage I: SAE Training (Reasoning Ability Extraction)}\; The first stage involves training an SAE to reconstruct the activations from a specific layer of a source model. We feed a trigger dataset, comprising only verified CoT-free question-answer pairs, into the source model and capture the resulting activations at a chosen SAE hookpoint. The SAE is then trained on these activations, learning features that represent the source model's internal reasoning while processing the trigger data. Our central insight is that a subset of these features corresponds to the source model's implicit reasoning abilities.
    \item \textbf{Stage II: SAE-Guided SFT (Reasoning Ability Elicitation)}\; Once the SAE has been trained, we shift to training a target model. The trained SAE is actively integrated into the target model's architecture (at certain layer) and kept with frozen weights during a standard SFT process. By exposing the target model to the feature representations captured by the SAE, the SFT process is guided to develop internal pathways that elicit reasoning abilities, effectively reconstructing such abilities extracted from the source model. This entire stage uses an elicitation dataset, which is typically identical to the trigger dataset.
\end{itemize}

\clearpage

The SAE-Tuning procedure is configured by five key components. Table \ref{tab:resa_lookup} details the configurations used to create each of our main Resa models in this paper. 

\begin{table}[!h]
\centering
\resizebox{\textwidth}{!}{
\begin{tabular}{l|ccccccc}
\toprule
\rowcolor{LightGreen}
\textbf{\textsc{Main Model}} & \textbf{\textsc{Source Model}} & \textbf{\textsc{Target Model}} & \textbf{\textsc{Trigger Data}} & \textbf{\textsc{Elicitation Data}} & \textbf{\textsc{SAE Training Mode}} \\
\midrule
\textbf{Resa-STILL-v1 (in Table~\ref{tab:reasoning replication})} & Tina-STILL & R1-Distill & STILL & STILL & Fine-tuned \\
\midrule
\textbf{Resa-STILL-v2 (in Table~\ref{tab:reasoning elicitation})} & - & R1-Distill & - & STILL & Pre-trained \\
\midrule
\textbf{Resa-STILL-v3 (in Table~\ref{tab:reasoning elicitation})} & Tina-STILL & R1-Distill & STILL & STILL & Trained-from-Scratch \\
\midrule
\textbf{Resa-STILL-v4 (in Table~\ref{tab:reasoning elicitation})} & R1-Distill & R1-Distill & STILL & STILL & Fine-tuned \\
\midrule
\textbf{Resa-STILL-v5 (in Table~\ref{tab:reasoning elicitation})} & \textbf{R1-Distill} & \textbf{R1-Distill} & \textbf{STILL} & \textbf{STILL} & \textbf{Trained-from-Scratch} \\
\midrule
\textbf{Resa-DeepScaleR-v1 (in Table~\ref{tab:reasoning replication})} & Tina-DeepScaleR & R1-Distill & DeepScaleR & DeepScaleR & Fine-tuned \\
\midrule
\textbf{Resa-DeepScaleR-v2 (in Table~\ref{tab:reasoning elicitation})} & Tina-DeepScaleR & R1-Distill & DeepScaleR & DeepScaleR & Trained-from-Scratch \\
\midrule
\textbf{Resa-DeepScaleR-v3 (in Table~\ref{tab:reasoning elicitation})} & \textbf{R1-Distill} & \textbf{R1-Distill} & \textbf{DeepScaleR} & \textbf{DeepScaleR} & \textbf{Trained-from-Scratch} \\
\midrule
\textbf{Resa-DeepScaleR-v4 (in Table~\ref{tab:universal and modular})} & Tina-STILL & R1-Distill & STILL & DeepScaleR & Fine-tuned \\
\bottomrule
\end{tabular}
}
\caption{\textbf{Configurations of Main Resa Models}\; Each row corresponds to a Resa model and outlines the components used in SAE-Tuning. The bolded rows represent a key configuration where reasoning abilities are extracted from the base R1-Distill model itself and then instilled back into the same model.}
\label{tab:resa_lookup}
\end{table}

In the following, we describe these components:
\begin{itemize}[leftmargin=7mm,itemsep=2mm, topsep=0em]
    \item \textbf{Source Model}\; The model from which reasoning-related features are extracted. The SAE is trained on the intermediate activations of one of its layers.
    \item \textbf{Target Model}\; The model for which we aim to elicit the reasoning abilities. In this paper, the target model is usually R1-Distill, which also serves as the base model of the source models, e.g., Tina models are more specialized models on R1-Distill.
    \item \textbf{Trigger Dataset}\; The CoT-free dataset used to trigger reasoning-related features in SAEs during SAE training. It is constructed from a standard open-source question-answer dataset by simply formatting each entry into a specific template. For a given question and its final answer, we have an input sequence:
    \begin{center}
        Problem: [Question] \texttt{<think>} [Answer] \texttt{</think>} \texttt{<answer>} Answer: [Answer] \texttt{</answer>}
    \end{center}
    Crucially, while this template uses \texttt{<think>} and \texttt{</think>} tokens, the dataset remains CoT-free as no intermediate reasoning steps are provided between the tokens, only the definitive answer is present. The inclusion of this structure is hypothesized to activate the source model's latent reasoning abilities, allowing the SAE to capture the corresponding features. A detailed analysis of the role and importance of these thinking tokens is provided in Section~\ref{sec:hypothesis transparent extraction}.
    \item \textbf{Elicitation Dataset}\; The CoT-free dataset used for the SAE-guided SFT of the target model to elicit reasoning abilities. In our experiments, it is usually the same as the trigger dataset.
    \item \textbf{SAE Training Mode}\; This defines how Stage I (i.e., SAE training) is carried out. We explore three distinct modes. (1) Pre-trained: We use the pre-trained SAE on R1-Distill from EleutherAI.\footnote{\href{https://huggingface.co/EleutherAI/sae-DeepSeek-R1-Distill-Qwen-1.5B-65k}{EleutherAI/sae-DeepSeek-R1-Distill-Qwen-1.5B-65k}} This mode bypasses Stage I entirely as there is no need to train the SAE. (2) Fine-tuned: The default pre-trained SAE is further fine-tuned on activations from the source model using the trigger dataset. (3) Trained-from-Scratch: An SAE is trained from a random initialization, exclusively on the activations produced by the source model with the trigger dataset.
\end{itemize}

\clearpage

In the following, We now formalize the two primary stages of the SAE-Tuning procedure.

\textbf{Stage I: SAE Training}\; Given a source model $M_{0}$, we denote the SAE to be trained as $s_{\ell}$, which is hooked at $\ell$-th layer (i.e., the multilayer perceptron output) of the source model $M_{0}.$
Suppose the source model has $L$ layers with hidden dimension $d$. Given input $\mathbf{x}_0$ and output $\mathbf{y}$, we denote the activation after the $\ell$th layer by $\mathbf{x}_{\ell}$. We view the $\ell$th transformer block as a function $h_{\ell}$ and we then have 
\begin{align}
    \mathbf{x}_{\ell} &= h_{\ell}(\mathbf{x}_{\ell-1}), \hspace{2mm} 1 \leq \ell \leq L, \\
    \mathbf{y} &= \text{softmax}(\mathbf{x}_L).
\end{align}
The SAE $s_{\ell}$ trains an encoder $\mathbf{W}_{\text{enc}} \in \mathbb{R}^{m \times d}$ for $m \gg d$, a decoder $\mathbf{W}_{\text{dec}} \in \mathbb{R}^{d \times m}$ with unit norm columns, and biases $\mathbf{b}_{\text{enc}} \in \mathbb{R}^m$, $\mathbf{b}_{\text{dec}} \in \mathbb{R}^d$. For activation $\mathbf{x}_{\ell}$, the SAE reconstructs activation $\tilde{\mathbf{x}}_{\ell}$ as
\begin{align}
    \mathbf{z} &= \text{Top-}k(\mathbf{W}_{\text{enc}} (\mathbf{x}_{\ell} - \mathbf{b}_{\text{dec}}) + \mathbf{b}_{\text{enc}}), \\
    \tilde{\mathbf{x}}_{\ell} &= \mathbf{W}_{\text{dec}} \mathbf{z} + \mathbf{b}_{\text{dec}} = \sum w_i f_i,
\end{align}
where Top-$k$ means that we only the top $k$ features in the vector~\citep{gao2024scalingevaluatingsparseautoencoders}. This is a simple and standard practice when training SAEs. 
During training, the SAE minimizes the reconstruction error
\begin{align}
\mathcal{L} = \|\mathbf{x}_{\ell} -\tilde{\mathbf{x}}_{\ell}\|^2.
\end{align}
\textbf{Stage II: SAE-Guided SFT}\; For the trained SAE $s_{\ell} $ with a hookpoint $\ell$, we freeze its weights and insert it\footnote{In this version, we only consider single-layer SAE insertion, i.e., insert at most one SAE at a time.} immediately after layer $\ell$ of a target model $M$. Note that this operation requires the source and target models to have the same underlying model architecture and size such that the SAE can be inserted directly. We denote the intermediate activation after $i$-th layer, before and after SAE insertion, as $\mathbf{x}_{i}$ and $\tilde{\mathbf{x}}_{i}$, respectively.
Given the SAE $s_{\ell}$ with a hookpoint $\ell$, we have $\tilde{\mathbf{x}}_{i} = \mathbf{x}_{i}, i \le \ell - 1$ and the reconstructed activation $\tilde{\mathbf{x}}_\ell = \text{SAE}(\mathbf{x}_\ell)$ propagates through the remaining layers to produce 
\begin{align}
    \tilde{\mathbf{x}}_i &= h_i(\tilde{\mathbf{x}}_{i-1}), \hspace{2mm} \ell + 1 \leq i \leq L, \\
    \tilde{\mathbf{y}} &= \text{softmax}(\tilde{\mathbf{x}}_{L}). 
\end{align}
We then add low-rank adapters of rank $r$ in each multilayer perceptron and attention sublayer of every layer of the target model we are adapting. We provide insights on why we choose low-rank adapters in Section~\ref{sec:hypothesis universal and modular}.
Concretely, for each frozen weight matrix $W_i \in \mathbb{R}^{d_1 \times d_2}$, we add $A_i \in \mathbb{R}^{d_1 \times r}$ and $B_i \in \mathbb{R}^{r \times d_2}$.
We train only the low-rank adapters $\Theta = \{A_i\} \cup \{B_i\}$. 
The training objective is the KL divergence between the next token probability distribution with and without the SAE inserted:
\begin{equation}
    \argmin_{\Theta} \mathcal{D}_{\text{KL}}(\tilde{\mathbf{y}}, \mathbf{y}).
\end{equation}
The core intuition behind this loss function is to force the target model to produce internal representations at $\ell$-th layer that are ``compatible'' with the frozen SAE with rich reasoning features. Since the SAE was trained to reconstruct the reasoning-focused activations of the source model, this objective pushes the target model's activations to become explicitly similar to the source model's internal reasoning structure. In essence, by fine-tuning the target model to accommodate the SAE with minimal disruption to its output, we are instilling the reasoning patterns embodied by the SAE's learned features.

Crucially, after this SAE-guided SFT is complete, the SAE is entirely removed from the model at test time. This leaves an enhanced target model with the elicited reasoning abilities ``instilled'' directly into its own parameters, ready for standard inference and evaluation.

\clearpage

In the following, we offer two more perspectives to build intuition for the SAE-Tuning procedure.
\begin{itemize}[leftmargin=7mm,itemsep=2mm, topsep=0em]
    \item \textbf{Perspective I: Knowledge Distillation}\; The SAE acts as a ``knowledge bottleneck.'' In Stage I, it is forced to learn a compressed and essential representation of the source model's reasoning processes. In Stage II, it becomes a ``teacher.'' The KL divergence objective distills this knowledge into the target (student) model's LoRA parameters, effectively teaching the student model to replicate the teacher's reasoning behavior while being guided by the explicit reasoning features captured by the SAE. We provide more detailed discussion of this perspective in Section~\ref{sec:practical utility}.
    \item \textbf{Perspective II: Alternating Optimization}\; We are optimizing two distinct sets of parameters for two different goals, one after the other. 1) Optimizing SAE: We hold the model constant and train the SAE parameters to best capture the model's latent reasoning features. 2) Optimizing Model Adapters: We hold the model and the SAE constant and train the low-rank adapters to best integrate the SAE's reasoning feature representation into the model. Combining these two goals allows the SAE to act as a bridge between the two models. The first optimization step builds this bridge by learning a compressed blueprint of the source model's reasoning process. The second step then fine-tunes the target model to align with and cross this bridge, ensuring it inherits the structural properties of the source model's reasoning without needing to learn them from scratch.
\end{itemize}

\subsection{Practical Implementation Setup}
\label{sec:implementation}

\begin{table}[!b]
\centering
\resizebox{.9\textwidth}{!}{
\begin{tabular}{l|ccccc}
\toprule
\rowcolor{LightGreen}
\textbf{\textsc{Experimental Task}} & \textbf{\textsc{Training Cost Est.}} & \textbf{\textsc{Evaluation Cost Est.}} & \textbf{\textsc{Total Cost Est.}} \\
\midrule
\textbf{Main: Resa-STILL-v1 (in Table~\ref{tab:reasoning replication})} & \$1 & \$4 & \$5 \\
\midrule
\textbf{Main: Resa-STILL-v2 (in Table~\ref{tab:reasoning elicitation})} & \$1 & \$4 & \$5 \\
\midrule
\textbf{Main: Resa-STILL-v3 (in Table~\ref{tab:reasoning elicitation})} & \$1 & \$4 & \$5 \\
\midrule
\textbf{Main: Resa-STILL-v4 (in Table~\ref{tab:reasoning elicitation})} & \$1 & \$4 & \$5 \\
\midrule
\textbf{Main: Resa-STILL-v5 (in Table~\ref{tab:reasoning elicitation})} & \$1 & \$4 & \$5 \\
\midrule
\midrule
\textbf{Main: Resa-DeepScaleR-v1 (in Table~\ref{tab:reasoning replication})} & \$1 & \$6 & \$7 \\
\midrule
\textbf{Main: Resa-DeepScaleR-v2 (in Table~\ref{tab:reasoning elicitation})} & \$1 & \$6 & \$7 \\
\midrule
\textbf{Main: Resa-DeepScaleR-v3 (in Table~\ref{tab:reasoning elicitation})} & \$1 & \$6 & \$7 \\
\midrule
\textbf{Main: Resa-DeepScaleR-v4 (in Table~\ref{tab:universal and modular})} & \$1 & \$6 & \$7 \\
\midrule
\midrule
\textbf{Ablation: Algorithm (in Table~\ref{tab:reasoning replication})} & \$2 & \$10 & \$12 \\
\midrule
\textbf{Ablation: Source Model (in Table~\ref{tab:reasoning elicitation})} & \$14 & \$60 & \$74 \\
\midrule
\textbf{Ablation: SAE Training Mode (in Table~\ref{tab:reasoning elicitation})} & \$3 & \$12 & \$15 \\
\midrule
\midrule
\textbf{Hypothesis: Generality (in Table~\ref{tab:universal and modular})} & \$7 & \$35 & \$42 \\
\midrule
\textbf{Hypothesis: Modularity (in Table~\ref{tab:universal and modular})} & \$102 & \$8 & \$120 \\
\midrule
\textbf{Hypothesis: Transparency (in Table~\ref{tab:layer ablation})} & \$26 & \$104 & \$130 \\
\midrule
\midrule
\textbf{Total: All Tasks} & \textbf{\$163} & \textbf{\$273} & \textbf{\$436} \\
\midrule
\textbf{Total: Main Tasks} & \textbf{\$9} & \textbf{\$44} & \textbf{\$53} \\
\midrule
\textbf{Total: Best Resa Model} & \textbf{\$1} & \textbf{\$4} & \textbf{\$5} \\
\bottomrule
\end{tabular}
}
\caption{\textbf{Computational Cost Breakdown}\; We provide a detailed cost breakdown of all experiments in this paper. Notice that the training cost estimate includes the costs for training both models and SAEs.} 
\label{tab:cost_breakdown}
\end{table}

In the following, we show our practical setup for implementing SAE-Tuning. Particularly, we demonstrate that this implementation is efficient and only requires minimal setup and computational cost.

\textbf{Training Setup}\; Our experiments are designed to be both effective and reproducible. The SAE-Tuning training code is built on the combination of \texttt{OpenR1}~\citep{openr1} and \texttt{Sparsify}.\footnote{\href{https://github.com/EleutherAI/sparsify}{https://github.com/EleutherAI/sparsify}} The default configuration for SAE-Tuning is as follows. The primary datasets are STILL\footnote{\href{https://huggingface.co/datasets/RUC-AIBOX/STILL-3-Preview-RL-Data}{RUC-AIBOX/STILL-3-Preview-RL-Data}}~\citep{Slow_Thinking_with_LLMs_3_Preview} and DeepScaleR\footnote{\href{https://huggingface.co/datasets/agentica-org/DeepScaleR-Preview-Dataset}{agentica-org/DeepScaleR-Preview-Dataset}}~\citep{deepscaler2025}. The source model varies across experiments, from specialized fine-tuned models (i.e., Tina models) to their base R1-Distill model~\citep{deepseekai2025deepseekr1incentivizingreasoningcapability}. By default, we choose the best Tina checkpoint trained on a specific dataset as the source model. The target model is by default the R1-Distill. For SAE, unless stated otherwise, we follow the ``fine-tuned'' training mode. The SAE is hooked to the output of the multilayer perceptron submodule after the $12$th layer (out of $28$) of the source model. This choice is based on the heuristic that middle layers in a transformer-based language model are often crucial for understanding and reasoning. We provide a detailed discussion on layer selection in Section~\ref{sec:hypothesis transparent extraction}. The full hyperparameter is provided in Appendix~\ref{app:hyperparameter}.

\textbf{Evaluation Setup}\; All evaluations reported herein utilize the \texttt{lighteval} framework~\citep{lighteval} integrated with the \texttt{vLLM}~\citep{kwon2023efficient} inference engine for efficiency. We maintain a fixed hardware configuration (two GPUs) and apply a standardized set of \texttt{vLLM} inference parameters across all evaluated models. All scores are zero-shot Pass@1 performance. Particularly, we evaluate the reasoning abilities of models across a diverse suite of six reasoning benchmarks, primarily focused on mathematical and scientific reasoning: AIME24/25~\citep{aime2025}, AMC23~\citep{amc23}, MATH500~\citep{hendrycks2021measuringmathematicalproblemsolving, lightman2023let}, Minerva~\citep{NEURIPS2022_18abbeef}, and GPQA Diamond (short as GPQA in this rest of the paper)~\citep{rein2024gpqa}.

\textbf{Overall Budget}\; A primary motivation for developing SAE-Tuning is to democratize research into reasoning models by establishing a low-cost and high-efficiency paradigm via SAEs. We deliberately constrain our setup to a minimal hardware footprint, using just 2 NVIDIA L40S or NVIDIA RTX 6000 Ada GPUs for all training and evaluation tasks. This setup is readily accessible on major cloud platforms, with an approximate cost of \$1 USD per GPU hour at the time of our experiments. As detailed in Table \ref{tab:cost_breakdown}, this approach demonstrates high cost-efficiency. We believe this setup provides a valuable testbed for the broader research community.

\section{Efficient Reasoning Ability Elicitation}
\label{sec:efficient elicitation}

In this section, we empirically validate the effectiveness of SAE-Tuning. We first demonstrate that it can successfully replicate the performance of models fully trained with RL. Building on this, we show its primary practical utility: that SAE-Tuning remains effective with the base R1-Distill model as source model, thus bypassing the need for expensive further RL. In addition, we establish the method's self-sufficiency with a trained-from-scratch SAE, which eliminates the dependence on pre-trained SAEs.

\subsection{Proof of Concept: Reasoning Ability Replication}
\label{sec:proof of concept}
We establish a proof of concept by answering: \textit{Can SAE-Tuning extract and transfer reasoning ability from a source model post-trained for reasoning via RL?} Therefore, we use the Tina models~\citep{wang2025tinatinyreasoningmodels}, which were trained with RL, as our source models. The goal is to see if our Resa models can match the Tina models' performance.

\begin{table}[!h]
\centering
\resizebox{.85\textwidth}{!}{
\begin{tabular}{l|ccccccc|cc}
\toprule
\rowcolor{LightGreen}
\textbf{\textsc{Model Name}} & \textbf{\textsc{AIME24}} & \textbf{\textsc{AIME25}} & \textbf{\textsc{AMC23}} & \textbf{\textsc{MATH500}} & \textbf{\textsc{GPQA}} & \textbf{\textsc{Minerva}} & \textbf{\textsc{Avg.}} \\
\midrule
\textbf{DeepSeek-R1-Distilled-Qwen-1.5B} & 23.33 & 16.67 & 62.50 & 82.60 & 31.82 & 30.15 & 41.18 \\
\midrule
\midrule
\textbf{STILL-3-1.5B-preview} & 26.67 & 26.67 & 67.50 & 86.40 & 34.34 & 27.57 & 44.86 \\
\midrule
\textbf{Tina-STILL (CoT-free RL)} & 36.67 & 30.00 & 77.50 & 84.60 & 33.33 & 26.84 & 48.16 \\
\midrule
\textbf{Resa-STILL-v1 (CoT-free SAE-Tuning)} & 33.33 & 33.33 & 75.00 & 83.80 & 29.41 & 28.79 & 47.28 \\
\midrule
\midrule
\textbf{DeepScaleR-1.5B-Preview} & 36.67 & 26.67 & 77.50 & 87.80 & 31.82 & 31.99 & 48.74 
\\
\midrule
\textbf{Tina-DeepScaleR (CoT-free RL)} & 43.33 & 26.67 & 67.50 & 86.20 & 37.88 & 28.68 & 48.38 \\
\midrule
\textbf{Resa-DeepScaleR-v1 (CoT-free SAE-Tuning)} & 36.67 & 23.33 & 85.00 & 83.00 & 32.35 & 33.33 & 48.95 \\
\midrule
\midrule
\rowcolor{LightGreen}
\textbf{\textsc{Algorithm Ablation}} & \textbf{\textsc{AIME24}} & \textbf{\textsc{AIME25}} & \textbf{\textsc{AMC23}} & \textbf{\textsc{MATH500}} & \textbf{\textsc{GPQA}} & \textbf{\textsc{Minerva}} & \textbf{\textsc{Avg.}} \\
\midrule
\textbf{STILL-CoT-free-SFT} & 20.00 & 16.67 & 60.00 & 81.20 & 29.78 & 26.36 & 39.00 \\
\midrule
\textbf{DeepScaleR-CoT-based-SFT} & 10.00 & 6.67 & 57.50 & 68.60 & 20.22 & 36.36 & 33.22 \\
\bottomrule
\end{tabular}
}
\caption{\textbf{Reasoning Ability Replication}\; SAE-Tuning successfully replicates the performance of RL-trained source models. Resa models (trained with SAE-Tuning on CoT-free data) achieve performance on par with or exceeding their Tina counterparts. More detailed results are shown in Appendix~\ref{app:tab replication}.}
\label{tab:reasoning replication}
\end{table}

Table~\ref{tab:reasoning replication} summarizes our proof of concept results. On the STILL dataset, our Resa-STILL-v1 (47.28\% avg) recovers 98.2\% of the performance of the RL-trained Tina-STILL (48.16\% avg). On the DeepScaleR dataset, our Resa-DeepScaleR-v1 (48.95\% avg) not only replicates but slightly surpasses the performance of its corresponding Tina-DeepScaleR source model (48.38\% avg). The algorithm ablation clearly demonstrates the necessity of our method using SAEs: standard SFT on the same CoT-free data (i.e., STILL-CoT-free-SFT) achieves a mere 39.00\% average, falling far short of both the Tina models and our Resa counterparts. This shows that simply training on the final answers is insufficient and the SAE-guided SFT is the critical ingredient for eliciting reasoning abilities. Furthermore, standard SFT on a CoT-based dataset\footnote{The DeepScaleR dataset originally contains CoT reasoning traces which are only used in this ablation study.} (i.e., DeepScaleR-CoT-based-SFT) also performs worse (33.22\% avg), suggesting that naive CoT-based training is not an effective strategy for improving reasoning and underscores the novelty of our CoT-free approach via SAE-Tuning.

\subsection{Practical Utility: End-to-End Reasoning Ability Elicitation}
\label{sec:practical utility}
Having proven that SAE-Tuning can replicate reasoning ability, we now study: \textit{Can one bypass the need for a specialized source model and the need for an existing SAE altogether?} Specifically, this section investigates if we can simplify SAE-Tuning to an end-to-end procedure such that one can elicit reasoning abilities directly from the base R1-Distill model, without the need for a pre-trained SAE for this model.

\textbf{SAE Simplification}\; The SAE training mode ablation in Table~\ref{tab:reasoning elicitation} shows that training an SAE from scratch on the trigger dataset (i.e, Resa-STILL-Trained-from-Scratch-SAE, 47.36\% avg) is just as effective as fine-tuning a generic pre-trained SAE on the same dataset (i.e., Resa-STILL-Finetuned-SAE, 47.28\% avg). Both outperform using a default, pre-trained SAE (44.99\% avg). The key insight is that the SAE's performance for reasoning ability elicitation hinges on its exposure to the specific reasoning features encapsulated in the data (i.e., the trigger dataset), while the general knowledge from its initial pre-training is less critical in SAE-Tuning. This result aligns with the knowledge distillation perspective of SAE-Tuning at the end of Section~\ref{sec:method}. In that view, the SAE is a ``teacher'' guiding the ``student'' (i.e., the target model). A static, pre-trained SAE is an ineffective teacher because it is ignorant of the ``curriculum''—the reasoning patterns in the trigger dataset. In contrast, both training from scratch and fine-tuning are effective because they ensure the teacher first learns the specific material it is meant to teach. Overall, the finding simplifies the pipeline by eliminating the need for a pre-trained SAE, which yields substantial compute savings by avoiding the costly pre-training on large corpora like SmolLM2~\citep{allal2025smollm2smolgoesbig} and RedPajama~\citep{weber2024redpajama}.

\begin{table}[!h]
\centering
\resizebox{.95\textwidth}{!}{
\begin{tabular}{l|ccccccc|cc}
\toprule
\rowcolor{LightGreen}
\textbf{\textsc{Model Name}} & \textbf{\textsc{AIME24}} & \textbf{\textsc{AIME25}} & \textbf{\textsc{AMC23}} & \textbf{\textsc{MATH500}} & \textbf{\textsc{GPQA}} & \textbf{\textsc{Minerva}} & \textbf{\textsc{Avg.}} \\
\midrule
\textbf{STILL-3-1.5B-preview} & 26.67 & 26.67 & 67.50 & 86.40 & 34.34 & 27.57 & 44.86 \\
\midrule
\textbf{Tina-STILL} & 36.67 & 30.00 & 77.50 & 84.60 & 33.33 & 26.84 & 48.16 \\
\midrule
\midrule
\rowcolor{LightGreen}
\textbf{\textsc{SAE Training Mode Ablation}} & \textbf{\textsc{AIME24}} & \textbf{\textsc{AIME25}} & \textbf{\textsc{AMC23}} & \textbf{\textsc{MATH500}} & \textbf{\textsc{GPQA}} & \textbf{\textsc{Minerva}} & \textbf{\textsc{Avg.}} \\
\midrule
\textbf{Resa-STILL-Finetuned-SAE (i.e., Resa-STILL-v1)} & 33.33 & 33.33 & 75.00 & 83.80 & 29.41 & 28.79 & 47.28 \\
\midrule
\textbf{Resa-STILL-Pretrained-SAE (i.e., Resa-STILL-v2)} & 23.33 & 23.33 & 72.50 & 85.40 & 30.51 & 34.85 & 44.99 \\
\midrule
\textbf{Resa-STILL-Trained-from-Scratch-SAE (i.e., Resa-STILL-v3)} & 33.33 & 33.33 & 70.00 & 83.00 & 30.15 & 34.34 & 47.36 \\
\midrule
\midrule
\rowcolor{LightGreen}
\textbf{\textsc{Source Model Ablation (Fine-tuned SAE)}} & \textbf{\textsc{AIME24}} & \textbf{\textsc{AIME25}} & \textbf{\textsc{AMC23}} & \textbf{\textsc{MATH500}} & \textbf{\textsc{GPQA}} & \textbf{\textsc{Minerva}} & \textbf{\textsc{Avg.}} \\
\midrule
\textbf{Resa-STILL-Tina-0-step (i.e., Resa-STILL-v4)} & 23.33 & 20.00 & 77.50 & 84.60 & 27.57 & 35.35 & 44.73 \\
\midrule
\textbf{Resa-STILL-Tina-1-step} & 40.00 & 26.67 & 70.00 & 83.20 & 31.62 & 36.36 & 47.98 \\
\midrule
\textbf{Resa-STILL-Tina-10-step} & 23.33 & 23.33 & 75.00 & 82.20 & 29.41 & 33.33 & 44.43 \\
\midrule
\textbf{Resa-STILL-Tina-50-step} & 33.33 & 26.67 & 72.50 & 82.60 & 27.57 & 38.89 & 46.93 \\
\midrule
\textbf{Resa-STILL-Tina-100-step} & 43.33 & 23.33 & 82.50 & 85.60 & 28.68 & 33.33 & 49.46 \\
\midrule
\textbf{Resa-STILL-Tina-500-step} & 36.67 & 26.67 & 80.00 & 83.80 & 31.62 & 31.82 & 48.43 \\
\midrule
\textbf{Resa-STILL-Best-Tina-2000-step (i.e., Resa-STILL-v1)} & 33.33 & 33.33 & 75.00 & 83.80 & 29.41 & 28.79 & 47.28 \\
\midrule
\textbf{Resa-STILL-Tina-3000-step} & 26.67 & 23.33 & 72.50 & 85.40 & 27.94 & 34.85 & 45.11 \\
\midrule
\midrule
\rowcolor{LightGreen}
\textbf{\textsc{Source Model Ablation (Trained-from-Scratch SAE)}} & \textbf{\textsc{AIME24}} & \textbf{\textsc{AIME25}} & \textbf{\textsc{AMC23}} & \textbf{\textsc{MATH500}} & \textbf{\textsc{GPQA}} & \textbf{\textsc{Minerva}} & \textbf{\textsc{Avg.}} \\
\midrule
\textbf{Resa-STILL-Tina-0-step (i.e., Resa-STILL-v5)} & 33.33 & 26.67 & 70.00 & 87.00 & 29.41 & 41.92 & 48.06 \\
\midrule
\textbf{Resa-STILL-Tina-1-step} & 33.33 & 33.33 & 72.50 & 82.20 & 29.41 & 35.35 & 47.69 \\
\midrule
\textbf{Resa-STILL-Tina-10-step} & 33.33 & 16.67 & 67.50 & 86.20 & 30.51 & 37.37 & 45.26 \\
\midrule
\textbf{Resa-STILL-Tina-50-step} & 43.33 & 23.33 & 77.50 & 83.40 & 29.41 & 38.89 & 49.31 \\
\midrule
\textbf{Resa-STILL-Tina-100-step} & 33.33 & 23.33 & 90.00 & 82.60 & 28.68 & 35.35 & 48.88 \\
\midrule
\textbf{Resa-STILL-Tina-500-step} & 36.67 & 20.00 & 67.50 & 84.20 & 30.88 & 35.35 & 45.77 \\
\midrule
\textbf{Resa-STILL-Best-Tina-2000-step (i.e., Resa-STILL-v3)} & 33.33 & 33.33 & 70.00 & 83.00 & 30.15 & 34.34 & 47.36 \\
\midrule
\textbf{Resa-STILL-Tina-3000-step} & 30.00 & 20.00 & 77.50 & 86.20 & 31.62 & 37.88 & 47.20 \\
\midrule
\midrule
\textbf{Resa-DeepScaleR-Best-Tina-1000-step (i.e., Resa-DeepScaleR-v2)} & 40.00 & 30.00 & 75.00 & 84.00 & 30.15 & 33.33 & 48.75 \\
\midrule
\textbf{Resa-DeepScaleR-Tina-0-step (i.e., Resa-DeepScaleR-v3)} & 33.33 & 23.33 & 80.00 & 86.00 & 30.51 & 31.31 & 47.41 \\
\bottomrule
\end{tabular}
}
\caption{\textbf{End-to-End Reasoning Ability Elicitation}\; (SAE Training Mode Ablation) Training an SAE from scratch is comparably as effective as fine-tuning a pre-trained SAE. (Source Model Ablations) The key results are Resa-STILL-v5 and Resa-DeepScaleR-v3 models which use the base model as its own source and match the reasoning performance of the RL-trained models. More detailed results are shown in Appendix~\ref{app:tab elicitation}.}
\label{tab:reasoning elicitation}
\end{table}

\textbf{Source Model Simplification}\; Based on the above SAE simplification, we then simplify the source model used for SAE training, using models ranging from the base R1-Distill model (i.e., Tina-0-step) to well-trained Tina checkpoints. We notice that the source of reasoning features is nuanced such that there is a non-monotonic relationship between the source model's training progression and the resulting Resa model's performance. The ``best'' reasoning features for extraction are not always found in the final, most-trained source model checkpoint. Specifically, it shows that one can get optimal performance with a light RL training, i.e., Resa-STILL-Tina-100-step (with fine-tuned SAE, 49.46\%) and Resa-STILL-Tina-50-step (with train-from-scratch SAE, 49.31\%). Another important finding is that by training an SAE from scratch and using the base model as the source, our method achieves a competitive average score of 48.06\% (i.e., Resa-STILL-v5). This performance is nearly identical to the fully RL-trained Tina-STILL model (48.16\%), demonstrating that our simplified, end-to-end SAE-Tuning procedure has potential to replace the RL fine-tuning stage with no meaningful loss in reasoning performance. This also confirms that the necessary reasoning features are already latent within the base model and can be elicited with high efficiency. Overall, this presents practitioners with a trade-off: using a lightly RL-trained source yields peak performance, while using the base model enables a maximally efficient, end-to-end workflow that still delivers competitive results.

\section{Hypothesis: Generalizable and Modular Reasoning Ability}
\label{sec:hypothesis universal and modular}

We now test a central hypothesis of our work: that the reasoning ability captured by SAE-Tuning is not a brittle, dataset-specific artifact but rather a generalizable and modular skill. We formulate this as a testable claim: \textit{Reasoning abilities extracted via SAEs can be transferred across both data distributions and models.} To validate this, we conduct two sets of experiments: First, we test out-of-distribution generalization by applying reasoning extracted from one dataset to another. Second, we test cross-model transfer by applying reasoning extracted from one model to another within the same family.

\begin{table}[!h]
\centering
\resizebox{.9\textwidth}{!}{
\begin{tabular}{l|ccccccc|cc}
\toprule
\rowcolor{LightGreen}
\textbf{\textsc{Out-of-Distribution Coverage Data}} & \textbf{\textsc{AIME24}} & \textbf{\textsc{AIME25}} & \textbf{\textsc{AMC23}} & \textbf{\textsc{MATH500}} & \textbf{\textsc{GPQA}} & \textbf{\textsc{Minerva}} & \textbf{\textsc{Avg.}} \\
\midrule
\textbf{DeepScaleR-1.5B-Preview} & 36.67 & 26.67 & 77.50 & 87.80 & 31.82 & 31.99 & 48.74 
\\
\midrule
\textbf{Tina-DeepScaleR} & 43.33 & 26.67 & 67.50 & 86.20 & 37.88 & 28.68 & 48.38 \\
\midrule
\textbf{Resa-STILL2DeepScaleR (i.e., Resa-DeepScaleR-v4)} & 33.33 & 30.00 & 80.00 & 84.00 & 29.41 & 35.86 & 48.77 \\
\midrule
\midrule
\rowcolor{LightGreen}
\textbf{\textsc{Out-of-Distribution Intersection Data}} & \textbf{\textsc{AIME24}} & \textbf{\textsc{AIME25}} & \textbf{\textsc{AMC23}} & \textbf{\textsc{MATH500}} & \textbf{\textsc{GPQA}} & \textbf{\textsc{Minerva}} & \textbf{\textsc{Avg.}} \\
\midrule
\textbf{Open-RS1} & 26.67 & 20.00 & 72.50 & 83.60 & 28.68 & 35.35 & 44.47 \\
\midrule
\textbf{Tina-Open-S1} & 43.33 & 20.00 & 80.00 & 84.00 & 28.68 & 35.35 & 48.56 \\
\midrule
\textbf{Resa-STILL2Open-S1} & 36.67 & 23.33 & 85.00 & 84.60 & 30.88 & 31.82 & 48.72 \\
\midrule
\midrule
\textbf{II-Thought-1.5B-Preview} & 30.00 & 23.33 & 72.50 & 86.80 & 30.88 & 31.90 & 45.90 \\
\midrule
\textbf{Tina-II-Thought} & 40.00 & 20.00 & 80.00 & 86.00 & 33.84 & 26.84 & 47.78 \\
\midrule
\textbf{Resa-STILL2II-Thought} & 40.00 & 23.33 & 75.00 & 83.20 & 31.25 & 38.89 & 48.61 \\
\midrule
\midrule
\textbf{Tina-OpenR1} & 36.67 & 26.67 & 75.00 & 86.80 & 30.51 & 39.90 & 49.26 \\
\midrule
\textbf{Resa-STILL2OpenR1} & 33.33 & 30.00 & 77.50 & 86.80 & 27.21 & 41.92 & 49.46 \\
\midrule
\midrule
\rowcolor{LightGreen}
\textbf{\textsc{Reasoning-as-an-Adapter}} & \textbf{\textsc{AIME24}} & \textbf{\textsc{AIME25}} & \textbf{\textsc{AMC23}} & \textbf{\textsc{MATH500}} & \textbf{\textsc{GPQA}} & \textbf{\textsc{Minerva}} & \textbf{\textsc{Avg.}} \\
\midrule
\textbf{Resa-STILL-Qwen-Math-Adapter} & 36.67 & 20.00 & 82.50 & 83.40 & 31.25 & 33.33 & 47.86 \\
\midrule
\textbf{Resa-STILL-Qwen-Adapter} & 30.00 & 30.00 & 72.50 & 85.60 & 31.25 & 35.86 & 47.54 \\
\bottomrule
\end{tabular}
}
\caption{\textbf{Generality and Modularity of Reasoning Ability}\; (Top \& middle) The results demonstrate OOD generalization across datasets. Resa-STILL2\textit{X} models are trained by extracting reasoning from the STILL dataset and applying it to a new elicitation dataset \textit{X}. (Bottom) The results demonstrate cross-model transfer. A reasoning adapter trained on Qwen-Math or Qwen is transferred to R1-Distill at inference time. More detailed results are shown in Appendix~\ref{app:tab universal and modular}.}
\label{tab:universal and modular}
\end{table}

\textbf{Out-of-Distribution Generalization}\; To assess out-of-distribution (OOD) generalization, we use a single dataset, STILL, to train the SAE on the source model (the ``trigger'' step). We then use that trained SAE to guide a SFT process of the target model on a completely different dataset (the ``elicit'' step). We test this on datasets that have varying degrees of overlap with STILL. Specifically, DeepScaleR fully covers the STILL dataset (which we refer as the coverage dataset) while Open-S1~\citep{dang2025reinforcementlearningreasoningsmall}, II-Thought~\citep{2025iithought}, and OpenR1~\citep{openr1} have underlying overlapped sources with STILL (which we coin as the intersection datasets). As shown in Table~\ref{tab:universal and modular}, the Resa-STILL2\textit{X} models, where reasoning ability from STILL is transferred to a new dataset \textit{X}, consistently achieve performance on par with models trained end-to-end via RL on that new dataset. For example, Resa-STILL2DeepScaleR scores 48.77\%, almost identical to Tina-DeepScaleR (48.38\%) which was trained entirely on DeepScaleR. This pattern holds across all tested datasets. This robust performance demonstrates that the reasoning features extracted from the STILL dataset are not overfitted to its specific data distribution. They represent a more general reasoning process that can be effectively applied to new distributions, showcasing OOD resilience.

\textbf{Modular Reasoning-as-an-Adapter}\; Recall from Section~\ref{sec:method} that during SAE-guided SFT, the parameters we train are all from low-rank adapters. Therefore, we explore if the extracted reasoning ability can similarly be treated as a modular ``adapter'' that can be plugged into other model. Specifically, we perform SAE-Tuning on models like Qwen-Math\footnote{\href{https://huggingface.co/Qwen/Qwen2.5-Math-1.5B}{Qwen/Qwen2.5-Math-1.5B}}~\citep{yang2024qwen25mathtechnicalreportmathematical} and Qwen\footnote{\href{https://huggingface.co/Qwen/Qwen2.5-1.5B}{Qwen/Qwen2.5-1.5B}}~\citep{qwen2025qwen25technicalreport} to produce a set of adapters. Then, at test time, we attach such adapters to R1-Distill in the same family, without any further training. The models in this family share an architecture but differ in their foundational knowledge: R1-Distill has the most general knowledge, Qwen-Math is specialized with math data, and Qwen is the most basic. This tests whether our extracted reasoning abilities can be separated from the foundational knowledge of the model it was trained on. As shown in the final rows of Table~\ref{tab:universal and modular}, the adapter trained on Qwen-Math or Qwen and attached to R1-Distill achieves an average score of 47.86\% or 47.54\%, respectively. This performance is competitive with models where the entire SAE-Tuning process was performed directly on R1-Distill (e.g., Resa-STILL-v1, 47.28\% avg). This result provides evidence that:
\begin{center}
    Strong Reasoning Model $\approx$ Abstract Reasoning Ability + Foundational Knowledge.
\end{center}
Our SAE-Tuning procedure aims to isolate the ``Abstract Reasoning Ability'' component into a portable adapter and the final performance is then a direct combination of this adapter with a model that possesses sufficient ``Foundational Knowledge.'' This opens up possibilities for creating highly capable and efficient models by composing reasoning abilities and foundational knowledge.

\section{Hypothesis: Transparent Reasoning Feature Extraction}
\label{sec:hypothesis transparent extraction}

A core claim of SAE-Tuning is that it provides a transparent approach to reasoning. Having demonstrated that it works, we now investigate how it works. We notice that the performance varies not only depending on the source model but also depending on the specific layer chosen of the source model for SAE training. This moves us beyond heuristics for SAE layer selection to a hypothesis: \textit{The suitability of a model layer for reasoning is predictable and correlated with the presence of quantifiable reasoning features.} To test this hypothesis, we introduce a novel prompt-only reasoning feature extraction method and use it to establish the underlying correlation between these features and reasoning performance.

\begin{figure}[!h]
    \centering
    \includegraphics[width=\linewidth]{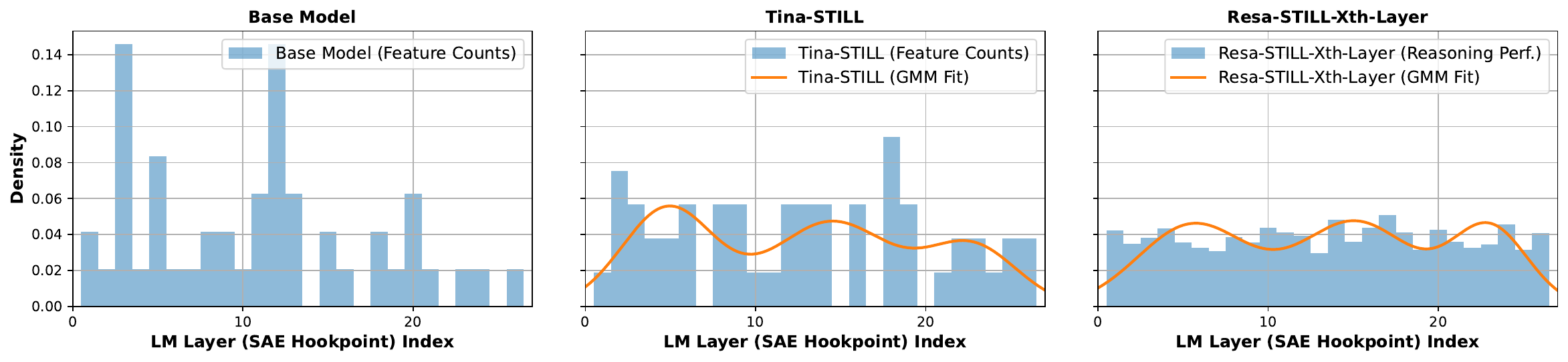}
    \caption{\textbf{Reasoning Feature Extraction}\; (Left) This shows the layer-wise feature counts of the base R1-Distill model. (Middle) This shows the layer-wise feature counts of the Tina-STILL model. (Right) This shows the reasoning performance of the trained Resa models with different layer-wise SAEs when Tina-STILL is the source model.}
    \label{fig:feature counts}
\end{figure}

\textbf{Prompt-Only Reasoning Feature Extraction}\; We propose a novel method to explicitly identify and quantify ``reasoning features'' and test if their distribution predicts the final performance of a Resa model. We hypothesize that features specifically involved in reasoning should activate primarily when the model is prompted to ``think.'' Specifically, we pass the standard DeepSeek-R1 system prompt containing \texttt{<think>} and \texttt{</think>} tokens through a model equipped with trained SAEs inserted after each layer indexed from $2$ to $27$. We cut off the first and final layer from the total $28$ layers since these two layers are mainly used for embedding and next token prediction, respectively. We then define reasoning features as those SAE features that are exclusively and simultaneously activated at the \texttt{<think>} and \texttt{</think>} tokens and not by other parts of the prompt. Applying this method to the base R1-Distill model revealed an interesting pattern that the layer-wise count of these reasoning features exhibits a \textit{tri-modal distribution} around layer indices $3, 12$, and $20$ as shown in Figure~\ref{fig:feature counts}.

\begin{table}[!h]
\centering
% \resizebox{.95\textwidth}{!}{
\resizebox{.85\textwidth}{!}{
\begin{tabular}{l|cccccccc|cc}
\toprule
\rowcolor{LightGreen}
\textbf{\textsc{Model Name}} & \textbf{\textsc{AIME24}} & \textbf{\textsc{AIME25}} & \textbf{\textsc{AMC23}} & \textbf{\textsc{MATH500}} & \textbf{\textsc{GPQA}} & \textbf{\textsc{Minerva}} & \textbf{\textsc{Avg.}} & \textbf{Feature Cnts.} \\
\midrule
\textbf{Resa-STILL-2nd-Layer} & 26.67 & 30.00 & 80.00 & 83.20 & 29.78 & 37.37 & 47.84 & 1 \\
\midrule
\textbf{Resa-STILL-3rd-Layer} & 26.67 & 36.67 & 70.00 & 83.40 & 28.31 & 33.84 & 46.48 & 4 \\
\midrule
\textbf{Resa-STILL-4th-Layer} & 33.33 & 20.00 & 80.00 & 83.80 & 28.68 & 36.87 & 47.11 & 3 \\
\midrule
\textbf{Resa-STILL-5th-Layer} & 40.00 & 23.33 & 70.00 & 83.20 & 26.84 & 44.95 & 48.05 & 2 \\
\midrule
\textbf{Resa-STILL-6th-Layer} & 33.33 & 23.33 & 72.50 & 83.40 & 31.62 & 35.35 & 46.59 & 2 \\
\midrule
\textbf{Resa-STILL-7th-Layer} & 26.67 & 26.67 & 77.50 & 81.60 & 27.94 & 35.86 & 46.04 & 3 \\
\midrule
\textbf{Resa-STILL-8th-Layer} & 20.00 & 23.33 & 82.50 & 85.20 & 29.78 & 33.33 & 45.69 & 0 \\
\midrule
\textbf{Resa-STILL-9th-Layer} & 36.67 & 20.00 & 80.00 & 84.20 & 27.57 & 34.34 & 47.13 & 3 \\
\midrule
\textbf{Resa-STILL-10th-Layer} & 36.67 & 23.33 & 67.50 & 84.80 & 29.78 & 37.37 & 46.58 & 3 \\
\midrule
\textbf{Resa-STILL-11th-Layer} & 26.67 & 36.67 & 77.50 & 83.80 & 31.25 & 32.83 & 48.12 & 1 \\
\midrule
\textbf{Resa-STILL-12th-Layer} & 30.00 & 26.67 & 77.50 & 84.60 & 31.99 & 34.85 & 47.60 & 1 \\
\midrule
\textbf{Resa-STILL-13th-Layer} & 33.33 & 33.33 & 75.00 & 83.80 & 29.41 & 28.79 & 47.28 & 3 \\
\midrule
\textbf{Resa-STILL-14th-Layer} & 33.33 & 20.00 & 77.50 & 83.60 & 28.68 & 29.80 & 45.48 & 3 \\
\midrule
\textbf{Resa-STILL-15th-Layer} & 36.67 & 23.33 & 80.00 & 84.80 & 30.15 & 38.89 & 48.97 & 3 \\
\midrule
\textbf{Resa-STILL-16th-Layer} & 30.00 & 20.00 & 80.00 & 85.00 & 30.15 & 34.85 & 46.67 & 0 \\
\midrule
\textbf{Resa-STILL-17th-Layer} & 43.33 & 16.67 & 77.50 & 83.80 & 30.88 & 36.36 & 48.09 & 3 \\
\midrule
\textbf{Resa-STILL-18th-Layer} & 43.33 & 20.00 & 77.50 & 84.00 & 33.82 & 37.88 & 49.42 & 0 \\
\midrule
\textbf{Resa-STILL-19th-Layer} & 33.33 & 30.00 & 72.50 & 84.20 & 29.04 & 36.87 & 47.66 & 5 \\
\midrule
\textbf{Resa-STILL-20th-Layer} & 33.33 & 23.33 & 75.00 & 85.00 & 29.04 & 29.29 & 45.83 & 3 \\
\midrule
\textbf{Resa-STILL-21st-Layer} & 30.00 & 33.33 & 75.00 & 84.60 & 27.57 & 36.87 & 47.90 & 0 \\
\midrule
\textbf{Resa-STILL-22nd-Layer} & 33.33 & 23.33 & 75.00 & 82.20 & 31.99 & 34.34 & 46.70 & 1 \\
\midrule
\textbf{Resa-STILL-23rd-Layer} & 30.00 & 20.00 & 80.00 & 82.00 & 29.04 & 35.35 & 46.07 & 2 \\
\midrule
\textbf{Resa-STILL-24th-Layer} & 36.67 & 20.00 & 77.50 & 83.80 & 30.51 & 29.80 & 46.38 & 2 \\
\midrule
\textbf{Resa-STILL-25th-Layer} & 40.00 & 30.00 & 70.00 & 84.20 & 28.68 & 37.88 & 48.46 & 1 \\
\midrule
\textbf{Resa-STILL-26th-Layer} & 36.67 & 20.00 & 65.00 & 85.60 & 30.88 & 36.87 & 45.84 & 2 \\
\midrule
\textbf{Resa-STILL-27th-Layer} & 30.00 & 36.67 & 67.50 & 83.00 & 27.94 & 40.40 & 47.59 & 2 \\
\bottomrule
\end{tabular}
}
\caption{\textbf{SAE Hookpoints Ablation}\; Performance evaluation of Resa-STILL models where SAE-Tuning is applied to each layer indexed from 2 to 27 individually. Feature Cnts.\ is the number of identified reasoning features in the corresponding layer of the Tina-STILL model. More detailed results are shown in Appendix~\ref{app:tab layer ablation}.}
\label{tab:layer ablation}
\end{table}

\textbf{Feature Counts v.s. Reasoning Performance Correlation}\; To test the hypothesis that this feature count distribution can predict reasoning performance, we conducted a large-scale study that we created $26$ different Resa-STILL models, with each one generated by applying SAE-Tuning to a different layer of Tina-STILL, from layer $2$ to $27$. The results in Table~\ref{tab:layer ablation} confirm that the choice of SAE hookpoint is critical. The average reasoning score fluctuates significantly, ranging from a low of 45.48\% (Layer $14$) to a high of 49.42\% (Layer $18$). Also, a naive interpretation, assuming more reasoning features equals better performance, is proven false. For instance, Layer $18$ yields the top performance (49.42\%) but has $0$ identified reasoning features, while Layer $19$ has the most features ($5$) but achieves a lower score (47.66\%). This validates our earlier finding that the source of reasoning features is nuanced. 

Such result indicates a more complex relationship between reasoning features and reasoning abilities exists. The key insight to discover such relationship comes from analyzing the overall distributions rather than single points. Just as the feature counts across layers form a tri-modal distribution, so does the final reasoning performance. Therefore, we fit a 3-component Gaussian Mixture Model (3-GMM) to both distributions: (1) the a priori reasoning feature counts from the base model, and (2) the final reasoning scores from our $26$ Resa models. The GMM analysis reveals an interesting and close structural alignment between the two GMM distributions. The means of the three Gaussian components for the feature count distribution are located near layers $4.9$, $14.5$, and $22.7$. The reasoning performance distribution's components cluster around nearly identical means at layers $5.6$, $15.1$, and $23.0$. This similarity extends to the component weights, which represent the proportion of layers belonging to each cluster. The feature distribution's weights (41\%, 37\%, 22\%) are closely mirrored by the reasoning performance distribution's weights (39\%, 37\%, 24\%). The overall spread of the two distributions, as measured by entropy, are nearly identical ($3.194$ for feature counts vs. $3.202$ for reasoning performance). This suggests that while a single layer's feature count may not be a good predictor, the overall structure of how reasoning is organized into three distinct layer-clusters within the model is a robust predictor of how performance will be distributed. In practice, one can therefore analyze the source model's feature distribution to strategically identify layer-clusters likely to yield high-performing models, providing a data-driven and transparent method for optimizing the SAE-Tuning process.

\section{Related Work}
\label{sec:related_work}

\textbf{Reinforcement Learning for Reasoning Ability Elicitation}\; The structure of reasoning tasks lends itself well to RL approaches, primarily because the final output's correctness provides a clear and verifiable reward signal. This feedback loop helps the model develop more robust reasoning strategies~\citep{shao2024deepseekmathpushinglimitsmathematical, deepseekai2025deepseekr1incentivizingreasoningcapability}. Recently, a growing body of work suggests that RL primarily elicits and amplifies reasoning capabilities already embedded within pretrained models, rather than installing them from scratch. Training dynamics analysis supports this ``elicitation hypothesis,'' showing that post-training largely surfaces latent abilities \citep{zhao2025echochamberrlposttraining}. The elicitation hypothesis is substantiated by several findings. For example, significant reasoning gains are achievable through minimal parameter updates that merely teach the model a new output format \citep{wang2025tinatinyreasoningmodels}, and even through one-shot RL with data selection \citep{wang2025reinforcementlearningreasoninglarge}. More surprisingly, studies have shown that RL can surface reasoning skills even with spurious or incorrect rewards \citep{shao2025spurious}, indicating that the primary mechanism is the surfacing of useful, pre-existing representations. We in this paper show that one can perform such elicitation in a much more efficient way by bypassing RL. 

\textbf{Sparse Autoencoders}\; Recent advances in SAEs have enabled new approaches for analyzing and steering neural network computations. Building on the original SAE architecture proposed by \citet{cunningham2023sparseautoencodershighlyinterpretable}, subsequent work from \citet{anthropic2023features} demonstrated how these sparse bottleneck networks can decompose transformer activations into human-interpretable features. The scaling properties of SAEs were systematically studied in \citet{anthropic2024scaling}, establishing practical guidelines for training SAEs across model sizes. Recent innovations have improved SAE training stability and feature quality: \citet{oneill2024sparseautoencodersenablescalable} introduced scalable and reliable circuit identification techniques. Concurrent work by \citet{chen2025lowrankadaptingmodelssparse} has focused on optimizing SAE computational efficiency and integration with modern transformer architectures. \citet{karvonen2025saebench} developed a comprehensive evaluation on components of SAE training. In the SAE-Tuning procedure, we leverage SAEs specifically to isolate and extract the latent features that underpin reasoning abilities of the model.

\textbf{Model Steering}\; The use of SAEs for model steering builds on earlier work in activation editing \citep{alain2018probe}. \citet{panickssery2024steeringllama2contrastive} first demonstrated that SAE features could be used for controlled behavior modification, while \citet{bayat2025steeringlargelanguagemodel} later developed more precise steering vectors through feature subspace analysis. Recent work by \citet{obrien2025steeringlanguagemodelrefusal} demonstrated improved safety properties through SAE-mediated interventions and \citet{gan2025tsv} showing SAEs' transferability of steering across modalities. Alternative to SAEs, activation differences \citep{li2023iti} and recursive feature machines \citep{beaglehole2025rfm} are also widely used for steering, and \citet{wu2025axbench} evaluated the concept steering abilities of these methods. Besides steering, \citet{chen2025lowrankadaptingmodelssparse} demonstrates the feasibility of adapting models to pretrained SAEs. Our proposed procedure, SAE-Tuning, beyond steering and adapting, fully leverages sparse autoencoders to identify, extract, and elicit latent reasoning abilities.
\section{Conclusion}
\label{sec:conclusion}

In this work, we confronted the challenge of eliciting reasoning abilities from language models in a three-birds-one-stone way that is \textit{effective}, \textit{efficient}, and \textit{transparent}. We moved beyond the prevailing paradigms of resource-intensive RL and quality-sensitive CoT-based SFT. Specifically, we introduced SAE-Tuning, a novel procedure that leverages SAEs to identify, extract, and elicit latent reasoning abilities using only CoT-free data. Our extensive experiments validated this approach on three key fronts. First, we demonstrated that SAE-Tuning is a performant and practical method, capable of not only replicating the performance of RL-trained models but, more importantly, of eliciting equivalent reasoning abilities directly from certain base models. Second, we established the surprising generality of these extracted abilities, demonstrating both their robustness to out-of-distribution data and also their modularity as portable ``reasoning adapters.'' Finally, we provided a new layer of transparency, showing that reasoning features are distributed in a predictable pattern across model layers and that this structure correlates with the reasoning performance.

\section{Acknowledgments}

We want to express our gratitude to the broader open-source community. This research was made possible by leveraging numerous publicly available resources, including training and evaluation framework, open datasets, accessible pre-trained language models, and the insights shared through technical reports. The computational resources required for the experiments described herein were partially provided by the Center for Advanced Research Computing (CARC) at the University of Southern California (USC). We are grateful for the support which enabled the training and evaluation of our models. J.A.\ was supported by the National Science Foundation Graduate Research Fellowship Program under Grant No.\ DGE-1842487. Any opinions, findings, and conclusions or recommendations expressed in this material are those of the authors and do not necessarily reflect the views of the National Science Foundation.

\clearpage
\bibliography{main}

\appendix
\clearpage

\appendix

\section*{\hspace{-4mm} \centering Appendix}
\vspace{3mm}

\section{Full Hyperparameter}
\label{app:hyperparameter}

We show our default choice of hyperparameter in Table~\ref{tab:common_hyperparameter}. The differences between main and ablation experiments largely lie in the hyperparameter we ablate over, which means the most of following hyperparameter is held constant across all experiments.

\begin{table}[h]
    \small
    \centering
    \begin{tabular}{lccccc}
        \toprule
        \midrule
        \rowcolor{LightGreen} SAE-Tuning Stage I: SAE Training & & & & & \\
        \midrule
        Number of features & \multicolumn{5}{c}{65536} \\
        Dead feature threshold & \multicolumn{5}{c}{1e6} \\
        Expansion factor & \multicolumn{5}{c}{64} \\
        Top-$k$ value & \multicolumn{5}{c}{32} \\
        Decoder normalization & \multicolumn{5}{c}{True} \\
        \midrule
        Optimizer & \multicolumn{5}{c}{Signum} \\
        Epochs & \multicolumn{5}{c}{1} \\
        Batch Size & \multicolumn{5}{c}{16} \\
        Learning Rate & \multicolumn{5}{c}{2.5e-4}\\
        Learning Rate Scheduler & \multicolumn{5}{c}{Constant} \\
        \midrule
        \midrule
        \rowcolor{LightGreen} SAE-Tuning Stage II: SAE-Guided SFT & & & & &  \\
        \midrule
        LoRA Modules & \multicolumn{5}{c}{query, key, value, dense} \\
        LoRA Rank & \multicolumn{5}{c}{32} \\
        LoRA $\alpha$ & \multicolumn{5}{c}{128} \\
        LoRA Dropout & \multicolumn{5}{c}{0.05} \\
        \midrule
        Optimizer & \multicolumn{5}{c}{AdamW} \\
        Optimizer Momentum & \multicolumn{5}{c}{$\beta_1$, $\beta_2$ = 0.9, 0.999} \\
        Epochs & \multicolumn{5}{c}{2} \\
        Batch Size & \multicolumn{5}{c}{1} \\
        Learning Rate & \multicolumn{5}{c}{1e-6}\\
        Learning Rate Scheduler & \multicolumn{5}{c}{Cosine with Min LR} \\
        \bottomrule
    \end{tabular}
    \vspace{3mm}
    \caption{\textbf{Default Hyperparameter Settings of SAE-Tuning} (Top) The default setting of SAE training. (Bottom) The default setting of SAE-Guided SFT.}
    \label{tab:common_hyperparameter}
\end{table}

\clearpage

\section{Additional Experiment Results}
\label{app:additional results}

\subsection{Full Results of Table~\ref{tab:reasoning replication}}
\label{app:tab replication}

In the following tables, we present the full performance evaluation results of models in Table~\ref{tab:reasoning replication}.

\begin{table}[h]
\centering
\resizebox{.85\textwidth}{!}{
\begin{tabular}{c|cccccccc|c}
\toprule
\rowcolor{LightGreen}
\textbf{\textsc{Checkpoint Steps}} & \textbf{\textsc{AIME24}} & \textbf{\textsc{AIME25}} & \textbf{\textsc{AMC23}} & \textbf{\textsc{MATH500}} & \textbf{\textsc{GPQA}} & \textbf{\textsc{Minerva}} & \textbf{\textsc{Avg.}} \\
\midrule
1000 & 20.00 & 33.33 & 75.00 & 82.60 & 29.78 & 33.33 & 45.67 \\
\midrule
1500 & 33.33 & 23.33 & 75.00 & 82.80 & 30.88 & 30.81 & 46.03 \\
\midrule
2000 & 33.33 & 33.33 & 75.00 & 83.80 & 29.41 & 28.79 & 47.28 \\
\midrule
2500 & 30.00 & 23.33 & 77.50 & 84.20 & 26.47 & 33.33 & 45.81 \\
\bottomrule
\end{tabular}
}
\caption{\textbf{Performance of Resa-STILL-v1}\; Each epoch contains 1448 Steps.}
\end{table}

\begin{table}[h]
\centering
\resizebox{.9\textwidth}{!}{
\begin{tabular}{c|cccccccc|c}
\toprule
\rowcolor{LightGreen}
\textbf{\textsc{Checkpoint Steps}} & \textbf{\textsc{AIME24}} & \textbf{\textsc{AIME25}} & \textbf{\textsc{AMC23}} & \textbf{\textsc{MATH500}} & \textbf{\textsc{GPQA}} & \textbf{\textsc{Minerva}} & \textbf{\textsc{Avg.}} \\
\midrule
500 & 30.00 & 30.00 & 67.50 & 84.00 & 28.68 & 32.32 & 45.42 \\
\midrule
1000 & 16.67 & 10.00 & 67.50 & 83.00 & 30.15 & 37.88 & 40.87 \\
\midrule
1500 & 23.33 & 20.00 & 70.00 & 84.40 & 27.57 & 37.88 & 43.86 \\
\midrule
2000 & 33.33 & 23.33 & 70.00 & 86.00 & 29.04 & 36.87 & 46.43 \\
\midrule
2500 & 36.67 & 23.33 & 85.00 & 83.00 & 32.35 & 33.33 & 48.95 \\
\midrule
3000 & 20.00 & 26.67 & 65.00 & 83.80 & 30.51 & 35.86 & 43.64 \\
\midrule
3500 & 23.33 & 23.33 & 70.00 & 81.80 & 30.15 & 36.36 & 44.16 \\
\midrule
4000 & 33.33 & 13.33 & 72.50 & 83.80 & 29.41 & 36.36 & 44.79 \\
\midrule
4500 & 36.67 & 13.33 & 70.00 & 83.80 & 27.94 & 31.31 & 43.84 \\
\midrule
5000 & 30.00 & 23.33 & 67.50 & 84.40 & 28.68 & 32.32 & 44.37 \\
\midrule
\bottomrule
\end{tabular}
}
\caption{\textbf{Performance of Resa-DeepScaleR-v1}\; Each epoch contains 2630 Steps.}
\end{table}

\pagebreak

\begin{table}[h]
\centering
\resizebox{.9\textwidth}{!}{
\begin{tabular}{c|cccccccc|c}
\toprule
\rowcolor{LightGreen}
\textbf{\textsc{Checkpoint Steps}} & \textbf{\textsc{AIME24}} & \textbf{\textsc{AIME25}} & \textbf{\textsc{AMC23}} & \textbf{\textsc{MATH500}} & \textbf{\textsc{GPQA}} & \textbf{\textsc{Minerva}} & \textbf{\textsc{Avg.}} \\
\midrule
500 & 20.00 & 16.67 & 60.00 & 81.20 & 29.78 & 26.36 & 39.00 \\
\midrule
1000 & 23.33 & 16.67 & 62.50 & 77.20 & 26.47 & 26.26 & 38.74 \\
\midrule
1500 & 13.33 & 10.00 & 60.00 & 74.00 & 28.68 & 27.78 & 35.63 \\
\midrule
\bottomrule
\end{tabular}
}
\caption{\textbf{Performance of STILL-CoT-free-SFT}\; Each epoch contains 936 Steps.}
\end{table}

\begin{table}[h]
\centering
\resizebox{.9\textwidth}{!}{
\begin{tabular}{c|cccccccc|c}
\toprule
\rowcolor{LightGreen}
\textbf{\textsc{Checkpoint Steps}} & \textbf{\textsc{AIME24}} & \textbf{\textsc{AIME25}} & \textbf{\textsc{AMC23}} & \textbf{\textsc{MATH500}} & \textbf{\textsc{GPQA}} & \textbf{\textsc{Minerva}} & \textbf{\textsc{Avg.}} \\
\midrule
1000 & 10.00 & 6.67 & 57.50 & 68.60 & 20.22 & 36.36 & 33.22 \\
\midrule
2000 & 16.67 & 6.67 & 52.50 & 67.40 & 21.69 & 28.28 & 32.20 \\
\midrule
3000 & 10.00 & 6.67 & 37.50 & 64.20 & 25.37 & 32.32 & 29.34 \\
\midrule
4000 & 10.00 & 6.67 & 35.00 & 61.80 & 22.79 & 27.78 & 27.34 \\
\midrule
5000 & 10.00 & 0.00 & 32.50 & 64.40 & 23.53 & 29.29 & 26.62 \\
\midrule
6000 & 0.00 & 6.67 & 40.00 & 64.00 & 23.53 & 28.79 & 27.16 \\
\midrule
7000 & 10.00 & 0.00 & 42.50 & 60.20 & 20.22 & 25.76 & 26.45 \\
\bottomrule
\end{tabular}
}
\caption{\textbf{Performance of DeepScaleR-CoT-based-SFT}\; Each epoch contains 2520 Steps.}
\end{table}

\clearpage
\subsection{Full Results of Table~\ref{tab:reasoning elicitation}}
\label{app:tab elicitation}

In the following tables, we present the full performance evaluation results of models in Table~\ref{tab:reasoning elicitation}.

\begin{table}[h]
\centering
\resizebox{.9\textwidth}{!}{
\begin{tabular}{c|cccccccc|c}
\toprule
\rowcolor{LightGreen}
\textbf{\textsc{Checkpoint Steps}} & \textbf{\textsc{AIME24}} & \textbf{\textsc{AIME25}} & \textbf{\textsc{AMC23}} & \textbf{\textsc{MATH500}} & \textbf{\textsc{GPQA}} & \textbf{\textsc{Minerva}} & \textbf{\textsc{Avg.}} \\
\midrule
1000 & 23.33 & 23.33 & 72.50 & 85.40 & 30.51 & 34.85 & 44.99 \\
\midrule
1500 & 20.00 & 20.00 & 75.00 & 81.40 & 29.78 & 30.30 & 42.75 \\
\midrule
2000 & 23.33 & 23.33 & 67.50 & 83.60 & 29.78 & 35.86 & 43.90 \\
\midrule
2500 & 23.33 & 26.67 & 67.50 & 83.00 & 25.74 & 34.34 & 43.43 \\
\bottomrule
\end{tabular}
}
\caption{\textbf{Performance of Resa-STILL-Pretrained-SAE}\; Each epoch contains 1448 Steps.}
\end{table}

\begin{table}[h]
\centering
\resizebox{.9\textwidth}{!}{
\begin{tabular}{c|cccccccc|c}
\toprule
\rowcolor{LightGreen}
\textbf{\textsc{Checkpoint Steps}} & \textbf{\textsc{AIME24}} & \textbf{\textsc{AIME25}} & \textbf{\textsc{AMC23}} & \textbf{\textsc{MATH500}} & \textbf{\textsc{GPQA}} & \textbf{\textsc{Minerva}} & \textbf{\textsc{Avg.}} \\
\midrule
1000 & 30.00 & 20.00 & 67.50 & 84.40 & 28.31 & 34.85 & 44.18 \\
\midrule
1500 & 33.33 & 33.33 & 70.00 & 83.00 & 30.15 & 34.34 & 47.36 \\
\midrule
2000 & 33.33 & 26.67 & 72.50 & 81.60 & 30.51 & 29.29 & 45.65 \\
\midrule
2500 & 30.00 & 23.33 & 70.00 & 85.60 & 32.35 & 33.33 & 45.77 \\
\bottomrule
\end{tabular}
}
\caption{\textbf{Performance of Resa-STILL-Trained-from-Scratch-SAE}\; Each epoch contains 1448 Steps.}
\end{table}

\begin{table}[h]
\centering
\resizebox{.9\textwidth}{!}{
\begin{tabular}{c|cccccccc|c}
\toprule
\rowcolor{LightGreen}
\textbf{\textsc{Checkpoint Steps}} & \textbf{\textsc{AIME24}} & \textbf{\textsc{AIME25}} & \textbf{\textsc{AMC23}} & \textbf{\textsc{MATH500}} & \textbf{\textsc{GPQA}} & \textbf{\textsc{Minerva}} & \textbf{\textsc{Avg.}} \\
\midrule
1000 & 23.33&	23.33&	77.50&	82.20&	28.31&	31.82&	44.42 \\
\midrule
1500 & 30.00&	16.67&	77.50&	80.40&	31.25&	29.80&	44.27 \\
\midrule
2000 & 33.33&	16.67&	65.00&	83.40&	27.21&	34.85&	43.41 \\
\midrule
2500 & 23.33&	20.00&	77.50&	84.60&	27.57&	35.35&	44.73 \\
\bottomrule
\end{tabular}
}
\caption{\textbf{Performance of Resa-STILL-Tina-0-step (Fine-tuned SAE)}\; Each epoch contains 1448 Steps.}
\end{table}

\begin{table}[h]
\centering
\resizebox{.9\textwidth}{!}{
\begin{tabular}{c|cccccccc|c}
\toprule
\rowcolor{LightGreen}
\textbf{\textsc{Checkpoint Steps}} & \textbf{\textsc{AIME24}} & \textbf{\textsc{AIME25}} & \textbf{\textsc{AMC23}} & \textbf{\textsc{MATH500}} & \textbf{\textsc{GPQA}} & \textbf{\textsc{Minerva}} & \textbf{\textsc{Avg.}} \\
\midrule
1000 & 26.67&	20.00&	67.50&	84.00&	29.04&	35.35&	43.76 \\
\midrule
1500 & 40.00&	26.67&	70.00&	83.20&	31.62&	36.36&	47.98 \\
\midrule
2000 & 26.67&	26.67&	65.00&	83.40&	27.94&	36.36&	44.34 \\
\midrule
2500 & 36.67&	23.33&	65.00&	84.00&	30.88&	36.36&	46.04 \\
\bottomrule
\end{tabular}
}
\caption{\textbf{Performance of Resa-STILL-Tina-1-step (Fine-tuned SAE)}\; Each epoch contains 1448 Steps.}
\end{table}

\begin{table}[h]
\centering
\resizebox{.9\textwidth}{!}{
\begin{tabular}{c|cccccccc|c}
\toprule
\rowcolor{LightGreen}
\textbf{\textsc{Checkpoint Steps}} & \textbf{\textsc{AIME24}} & \textbf{\textsc{AIME25}} & \textbf{\textsc{AMC23}} & \textbf{\textsc{MATH500}} & \textbf{\textsc{GPQA}} & \textbf{\textsc{Minerva}} & \textbf{\textsc{Avg.}} \\
\midrule
1000 & 30.00&	16.67&	67.50&	83.80&	31.62&	32.32&	43.65 \\
\midrule
1500 & 23.33&	23.33&	75.00&	82.20&	29.41&	33.33&	44.43 \\
\midrule
2000 & 23.33&	23.33&	62.50&	84.20&	27.57&	34.85&	42.63 \\
\midrule
2500 & 20.00&	20.00&	70.00&	84.20&	27.94&	35.86&	43.00 \\
\bottomrule
\end{tabular}
}
\caption{\textbf{Performance of Resa-STILL-Tina-10-step (Fine-tuned SAE)}\; Each epoch contains 1448 Steps.}
\end{table}

\begin{table}[h]
\centering
\resizebox{.9\textwidth}{!}{
\begin{tabular}{c|cccccccc|c}
\toprule
\rowcolor{LightGreen}
\textbf{\textsc{Checkpoint Steps}} & \textbf{\textsc{AIME24}} & \textbf{\textsc{AIME25}} & \textbf{\textsc{AMC23}} & \textbf{\textsc{MATH500}} & \textbf{\textsc{GPQA}} & \textbf{\textsc{Minerva}} & \textbf{\textsc{Avg.}} \\
\midrule
1000 & 26.67&	23.33&	70.00&	83.00&	28.68&	38.89&	45.09 \\
\midrule
1500 & 20.00&	26.67&	65.00&	84.60&	30.51&	34.85&	43.60 \\
\midrule
2000 & 30.00&	33.33&	67.50&	82.80&	30.15&	36.36&	46.69 \\
\midrule
2500 & 33.33&	26.67&	72.50&	82.60&	27.57&	38.89&	46.93 \\
\bottomrule
\end{tabular}
}
\caption{\textbf{Performance of Resa-STILL-Tina-50-step (Fine-tuned SAE)}\; Each epoch contains 1448 Steps.}
\end{table}

\begin{table}[h]
\centering
\resizebox{.9\textwidth}{!}{
\begin{tabular}{c|cccccccc|c}
\toprule
\rowcolor{LightGreen}
\textbf{\textsc{Checkpoint Steps}} & \textbf{\textsc{AIME24}} & \textbf{\textsc{AIME25}} & \textbf{\textsc{AMC23}} & \textbf{\textsc{MATH500}} & \textbf{\textsc{GPQA}} & \textbf{\textsc{Minerva}} & \textbf{\textsc{Avg.}} \\
\midrule
1000 & 43.33&	23.33&	82.50&	85.60&	28.68&	33.33&	49.46 \\
\midrule
1500 & 40.00&	23.33&	72.50&	84.60&	25.74&	32.32&	46.42 \\
\midrule
2000 & 30.00&	16.67&	70.00&	85.40&	33.09&	33.33&	44.75 \\
\midrule
2500 & 23.33&	23.33&	72.50&	84.40&	30.15&	36.36&	45.01 \\
\bottomrule
\end{tabular}
}
\caption{\textbf{Performance of Resa-STILL-Tina-100-step (Fine-tuned SAE)}\; Each epoch contains 1448 Steps.}
\end{table}

\begin{table}[h]
\centering
\resizebox{.9\textwidth}{!}{
\begin{tabular}{c|cccccccc|c}
\toprule
\rowcolor{LightGreen}
\textbf{\textsc{Checkpoint Steps}} & \textbf{\textsc{AIME24}} & \textbf{\textsc{AIME25}} & \textbf{\textsc{AMC23}} & \textbf{\textsc{MATH500}} & \textbf{\textsc{GPQA}} & \textbf{\textsc{Minerva}} & \textbf{\textsc{Avg.}} \\
\midrule
1000 & 36.67&	26.67&	80.00&	83.80&	31.62&	31.82&	48.43 \\
\midrule
1500 & 30.00&	23.33&	82.50&	84.20&	31.62&	36.36&	48.00 \\
\midrule
2000 & 36.67&	26.67&	70.00&	85.60&	31.99&	38.89&	48.30 \\
\midrule
2500 & 26.67&	20.00&	72.50&	82.20&	28.31&	35.86&	44.26 \\
\bottomrule
\end{tabular}
}
\caption{\textbf{Performance of Resa-STILL-Tina-500-step (Fine-tuned SAE)}\; Each epoch contains 1448 Steps.}
\end{table}

\begin{table}[h]
\centering
\resizebox{.9\textwidth}{!}{
\begin{tabular}{c|cccccccc|c}
\toprule
\rowcolor{LightGreen}
\textbf{\textsc{Checkpoint Steps}} & \textbf{\textsc{AIME24}} & \textbf{\textsc{AIME25}} & \textbf{\textsc{AMC23}} & \textbf{\textsc{MATH500}} & \textbf{\textsc{GPQA}} & \textbf{\textsc{Minerva}} & \textbf{\textsc{Avg.}} \\
\midrule
1000 & 30.00&	20.00&	75.00&	83.00&	26.84&	32.32&	44.53 \\
\midrule
1500 & 26.67&	23.33&	72.50&	85.40&	27.94&	34.85&	45.11 \\
\midrule
2000 & 26.67&	26.67&	72.50&	83.60&	27.57&	30.81&	44.64 \\
\midrule
2500 & 23.33&	26.67&	67.50&	82.60&	30.51&	36.36&	44.49 \\
\bottomrule
\end{tabular}
}
\caption{\textbf{Performance of Resa-STILL-Tina-3000-step (Fine-tuned SAE)}\; Each epoch contains 1448 Steps.}
\end{table}

\begin{table}[h]
\centering
\resizebox{.9\textwidth}{!}{
\begin{tabular}{c|cccccccc|c}
\toprule
\rowcolor{LightGreen}
\textbf{\textsc{Checkpoint Steps}} & \textbf{\textsc{AIME24}} & \textbf{\textsc{AIME25}} & \textbf{\textsc{AMC23}} & \textbf{\textsc{MATH500}} & \textbf{\textsc{GPQA}} & \textbf{\textsc{Minerva}} & \textbf{\textsc{Avg.}} \\
\midrule
1000 & 33.33&	26.67&	70.00&	87.00&	29.41&	41.92&	48.06 \\
\midrule
1500 & 23.33&	23.33&	70.00&	84.00&	29.78&	34.85&	44.22 \\
\midrule
2000 & 36.67&	23.33&	72.50&	83.40&	28.68&	36.87&	46.91 \\
\midrule
2500 & 43.33&	23.33&	65.00&	84.20&	27.94&	29.80&	45.60 \\
\bottomrule
\end{tabular}
}
\caption{\textbf{Performance of Resa-STILL-Tina-0-step (Trained-from-Scratch SAE)}\; Each epoch contains 1448 Steps.}
\end{table}

\begin{table}[h]
\centering
\resizebox{.9\textwidth}{!}{
\begin{tabular}{c|cccccccc|c}
\toprule
\rowcolor{LightGreen}
\textbf{\textsc{Checkpoint Steps}} & \textbf{\textsc{AIME24}} & \textbf{\textsc{AIME25}} & \textbf{\textsc{AMC23}} & \textbf{\textsc{MATH500}} & \textbf{\textsc{GPQA}} & \textbf{\textsc{Minerva}} & \textbf{\textsc{Avg.}} \\
\midrule
1000 & 30.00&	23.33&	75.00&	83.80&	25.74&	34.85&	45.45 \\
\midrule
1500 & 26.67&	23.33&	72.50&	85.00&	29.78&	38.38&	45.94 \\
\midrule
2000 & 33.33&	23.33&	65.00&	82.00&	29.78&	34.85&	44.72 \\
\midrule
2500 & 33.33&	33.33&	72.50&	82.20&	29.41&	35.35&	47.69 \\
\bottomrule
\end{tabular}
}
\caption{\textbf{Performance of Resa-STILL-Tina-1-step (Trained-from-Scratch SAE)}\; Each epoch contains 1448 Steps.}
\end{table}

\begin{table}[h]
\centering
\resizebox{.9\textwidth}{!}{
\begin{tabular}{c|cccccccc|c}
\toprule
\rowcolor{LightGreen}
\textbf{\textsc{Checkpoint Steps}} & \textbf{\textsc{AIME24}} & \textbf{\textsc{AIME25}} & \textbf{\textsc{AMC23}} & \textbf{\textsc{MATH500}} & \textbf{\textsc{GPQA}} & \textbf{\textsc{Minerva}} & \textbf{\textsc{Avg.}} \\
\midrule
1000 & 26.67&	20.00&	67.50&	85.80&	28.31&	36.87&	44.19 \\
\midrule
1500 & 33.33&	16.67&	67.50&	86.20&	30.51&	37.37&	45.26 \\
\midrule
2000 & 33.33&	23.33&	67.50&	84.40&	29.41&	32.32&	45.05 \\
\midrule
2500 & 26.67&	30.00&	62.50&	82.20&	26.84&	34.85&	43.84 \\
\bottomrule
\end{tabular}
}
\caption{\textbf{Performance of Resa-STILL-Tina-10-step (Trained-from-Scratch SAE)}\; Each epoch contains 1448 Steps.}
\end{table}

\begin{table}[h]
\centering
\resizebox{.9\textwidth}{!}{
\begin{tabular}{c|cccccccc|c}
\toprule
\rowcolor{LightGreen}
\textbf{\textsc{Checkpoint Steps}} & \textbf{\textsc{AIME24}} & \textbf{\textsc{AIME25}} & \textbf{\textsc{AMC23}} & \textbf{\textsc{MATH500}} & \textbf{\textsc{GPQA}} & \textbf{\textsc{Minerva}} & \textbf{\textsc{Avg.}} \\
\midrule
1000 & 36.67&	23.33&	80.00&	84.20&	28.68&	35.86&	48.12 \\
\midrule
1500 & 23.33&	33.33&	72.50&	84.00&	27.21&	35.86&	46.04 \\
\midrule
2000 & 30.00&	30.00&	70.00&	83.20&	32.72&	37.37&	47.22 \\
\midrule
2500 & 43.33&	23.33&	77.50&	83.40&	29.41&	38.89&	49.31 \\
\bottomrule
\end{tabular}
}
\caption{\textbf{Performance of Resa-STILL-Tina-50-step (Trained-from-Scratch SAE)}\; Each epoch contains 1448 Steps.}
\end{table}

\begin{table}[h]
\centering
\resizebox{.9\textwidth}{!}{
\begin{tabular}{c|cccccccc|c}
\toprule
\rowcolor{LightGreen}
\textbf{\textsc{Checkpoint Steps}} & \textbf{\textsc{AIME24}} & \textbf{\textsc{AIME25}} & \textbf{\textsc{AMC23}} & \textbf{\textsc{MATH500}} & \textbf{\textsc{GPQA}} & \textbf{\textsc{Minerva}} & \textbf{\textsc{Avg.}} \\
\midrule
1000 & 33.33&	23.33&	90.00&	82.60&	28.68&	35.35&	48.88 \\
\midrule
1500 & 46.67&	20.00&	62.50&	83.00&	28.31&	31.31&	45.30 \\
\midrule
2000 & 36.67&	20.00&	75.00&	83.20&	30.51&	38.38&	47.29 \\
\midrule
2500 & 30.00&	23.33&	65.00&	83.20&	29.04&	35.35&	44.32 \\
\bottomrule
\end{tabular}
}
\caption{\textbf{Performance of Resa-STILL-Tina-100-step (Trained-from-Scratch SAE)}\; Each epoch contains 1448 Steps.}
\end{table}

\begin{table}[h]
\centering
\resizebox{.9\textwidth}{!}{
\begin{tabular}{c|cccccccc|c}
\toprule
\rowcolor{LightGreen}
\textbf{\textsc{Checkpoint Steps}} & \textbf{\textsc{AIME24}} & \textbf{\textsc{AIME25}} & \textbf{\textsc{AMC23}} & \textbf{\textsc{MATH500}} & \textbf{\textsc{GPQA}} & \textbf{\textsc{Minerva}} & \textbf{\textsc{Avg.}} \\
\midrule
1000 & 23.33&	20.00&	75.00&	84.60&	30.51&	33.84&	44.55 \\
\midrule
1500 & 30.00&	20.00&	70.00&	83.80&	30.51&	32.32&	44.44 \\
\midrule
2000 & 36.67&	20.00&	67.50&	84.20&	30.88&	35.35&	45.77 \\
\midrule
2500 & 20.00&	23.33&	67.50&	82.80&	28.68&	35.35&	42.94 \\
\bottomrule
\end{tabular}
}
\caption{\textbf{Performance of Resa-STILL-Tina-500-step (Trained-from-Scratch SAE)}\; Each epoch contains 1448 Steps.}
\end{table}

\begin{table}[h]
\centering
\resizebox{.9\textwidth}{!}{
\begin{tabular}{c|cccccccc|c}
\toprule
\rowcolor{LightGreen}
\textbf{\textsc{Checkpoint Steps}} & \textbf{\textsc{AIME24}} & \textbf{\textsc{AIME25}} & \textbf{\textsc{AMC23}} & \textbf{\textsc{MATH500}} & \textbf{\textsc{GPQA}} & \textbf{\textsc{Minerva}} & \textbf{\textsc{Avg.}} \\
\midrule
1000 & 16.67&	23.33&	65.00&	83.60&	30.88&	34.85&	42.39 \\
\midrule
1500 & 30.00&	20.00&	77.50&	86.20&	31.62&	37.88&	47.20 \\
\midrule
2000 & 33.33&	26.67&	65.00&	85.20&	31.62&	35.35&	46.20 \\
\midrule
2500 & 20.00&	26.67&	67.50&	82.40&	28.68&	32.83&	43.01 \\
\bottomrule
\end{tabular}
}
\caption{\textbf{Performance of Resa-STILL-Tina-3000-step (Trained-from-Scratch SAE)}\; Each epoch contains 1448 Steps.}
\end{table}

\clearpage

\begin{table}[h]
\centering
\resizebox{.9\textwidth}{!}{
\begin{tabular}{c|cccccccc|c}
\toprule
\rowcolor{LightGreen}
\textbf{\textsc{Checkpoint Steps}} & \textbf{\textsc{AIME24}} & \textbf{\textsc{AIME25}} & \textbf{\textsc{AMC23}} & \textbf{\textsc{MATH500}} & \textbf{\textsc{GPQA}} & \textbf{\textsc{Minerva}} & \textbf{\textsc{Avg.}} \\
\midrule
1000 & 30.00&	23.33&	70.00&	86.40&	27.94&	32.83&	45.08 \\
\midrule
1500 & 33.33&	23.33&	80.00&	86.00&	30.51&	31.31&	47.41 \\
\midrule
2000 & 30.00&	16.67&	77.50&	83.60&	28.31&	31.82&	44.65 \\
\midrule
2500 & 23.33&	20.00&	75.00&	82.00&	29.78&	36.36&	44.41 \\
\midrule
3000 & 26.67&	16.67&	72.50&	83.20&	31.62&	33.33&	44.00 \\
\midrule
3500 & 30.00&	23.33&	75.00&	85.80&	27.94&	30.80&	45.48 \\
\bottomrule
\end{tabular}
}
\caption{\textbf{Performance of Resa-DeepScaleR-Best-Tina-1000-step (Trained-from-Scratch SAE)}\; Each epoch contains 1914 Steps.}
\end{table}

\begin{table}[h]
\centering
\resizebox{.9\textwidth}{!}{
\begin{tabular}{c|cccccccc|c}
\toprule
\rowcolor{LightGreen}
\textbf{\textsc{Checkpoint Steps}} & \textbf{\textsc{AIME24}} & \textbf{\textsc{AIME25}} & \textbf{\textsc{AMC23}} & \textbf{\textsc{MATH500}} & \textbf{\textsc{GPQA}} & \textbf{\textsc{Minerva}} & \textbf{\textsc{Avg.}} \\
\midrule
1000 & 26.67&	16.67&	75.00&	84.60&	28.68&	31.80&	43.90 \\
\midrule
1500 & 36.67&	30.00&	77.50&	83.80&	29.41&	31.26&	48.11 \\
\midrule
2000 & 23.33&	20.00&	82.50&	82.40&	28.68&	34.36&	45.21 \\
\midrule
2500 & 33.33&	23.33&	67.50&	83.00&	29.04&	32.43&	44.77 \\
\midrule
3000 & 40.00&	20.00&	72.50&	84.40&	26.47&	35.35&	46.45 \\
\midrule
3500 & 40.00&	30.00&	75.00&	84.00&	30.15&	33.33&	48.75 \\
\bottomrule
\end{tabular}
}
\caption{\textbf{Performance of Resa-DeepScaleR-Tina-0-step (Trained-from-Scratch SAE)}\; Each epoch contains 1914 Steps.}
\end{table}

\clearpage
\subsection{Full Results of Table~\ref{tab:universal and modular}}
\label{app:tab universal and modular}

In the following tables, we present the full performance evaluation results of models in Table~\ref{tab:universal and modular}.

\begin{table}[h]
\centering
\resizebox{.9\textwidth}{!}{
\begin{tabular}{c|cccccccc|c}
\toprule
\rowcolor{LightGreen}
\textbf{\textsc{Checkpoint Steps}} & \textbf{\textsc{AIME24}} & \textbf{\textsc{AIME25}} & \textbf{\textsc{AMC23}} & \textbf{\textsc{MATH500}} & \textbf{\textsc{GPQA}} & \textbf{\textsc{Minerva}} & \textbf{\textsc{Avg.}} \\
\midrule
500 & 33.33&	30.00&	80.00&	84.00&	29.41&	35.86&	48.77 \\
\midrule
1000 & 33.33&	23.33&	70.00&	84.80&	29.41&	32.83&	45.62 \\
\midrule
1500 & 26.67&	13.33&	67.50&	83.80&	31.25&	36.87&	43.24 \\
\midrule
2000 & 26.67&	23.33&	70.00&	83.20&	28.68&	34.34&	44.37 \\
\midrule
2500 & 33.33&	16.67&	70.00&	82.40&	27.57&	35.35&	44.22 \\
\midrule
3000 & 30.00&	26.67&	57.50&	83.20&	28.68&	37.37&	43.90 \\
\midrule
3500 & 26.67&	13.33&	77.50&	83.20&	28.31&	34.85&	43.98 \\
\midrule
4000 & 30.00&	23.33&	60.00&	84.40&	26.84&	37.37&	43.66 \\
\midrule
4500 & 36.67&	20.00&	75.00&	83.80&	27.94&	37.37&	46.80 \\
\midrule
5000 & 36.67&	20.00&	72.50&	84.00&	26.84&	31.82&	45.31 \\
\bottomrule
\end{tabular}
}
\caption{\textbf{Performance of Resa-STILL2DeepScaleR}\; Each epoch contains 1914 Steps.}
\end{table}

\begin{table}[h]
\centering
\resizebox{.9\textwidth}{!}{
\begin{tabular}{c|cccccccc|c}
\toprule
\rowcolor{LightGreen}
\textbf{\textsc{Checkpoint Steps}} & \textbf{\textsc{AIME24}} & \textbf{\textsc{AIME25}} & \textbf{\textsc{AMC23}} & \textbf{\textsc{MATH500}} & \textbf{\textsc{GPQA}} & \textbf{\textsc{Minerva}} & \textbf{\textsc{Avg.}} \\
\midrule
500 & 23.33&	20.00&	72.50&	84.00&	30.51&	32.83&	43.86 \\
\midrule
1000 & 36.67&	23.33&	85.00&	84.60&	30.88&	31.82&	48.72 \\
\midrule
1500 & 33.33&	26.67&	72.50&	83.40&	29.41&	39.90&	47.54 \\
\midrule
2000 & 30.00&	26.67&	75.00&	84.00&	30.88&	37.88&	47.41 \\
\bottomrule
\end{tabular}
}
\caption{\textbf{Performance of Resa-STILL2Open-S1}\; Each epoch contains 1063 Steps.}
\end{table}

\begin{table}[h]
\centering
\resizebox{.9\textwidth}{!}{
\begin{tabular}{c|cccccccc|c}
\toprule
\rowcolor{LightGreen}
\textbf{\textsc{Checkpoint Steps}} & \textbf{\textsc{AIME24}} & \textbf{\textsc{AIME25}} & \textbf{\textsc{AMC23}} & \textbf{\textsc{MATH500}} & \textbf{\textsc{GPQA}} & \textbf{\textsc{Minerva}} & \textbf{\textsc{Avg.}} \\
\midrule
1000 & 30.00&	20.00&	75.00&	84.60&	27.21&	34.34&	45.19 \\
\midrule
2000 & 40.00&	23.33&	75.00&	83.20&	31.25&	38.89&	48.61 \\
\midrule
3000 & 26.67&	23.33&	75.00&	85.20&	28.31&	36.36&	45.81 \\
\midrule
4000 & 26.67&	13.33&	67.50&	85.80&	27.94&	41.41&	43.78 \\
\midrule
5000 & 26.67&	20.00&	72.50&	85.60&	29.41&	37.37&	45.26 \\
\bottomrule
\end{tabular}
}
\caption{\textbf{Performance of Resa-STILL2II-Thought}\; Each epoch contains 2664 Steps.}
\end{table}

\begin{table}[h]
\centering
\resizebox{.9\textwidth}{!}{
\begin{tabular}{c|cccccccc|c}
\toprule
\rowcolor{LightGreen}
\textbf{\textsc{Checkpoint Steps}} & \textbf{\textsc{AIME24}} & \textbf{\textsc{AIME25}} & \textbf{\textsc{AMC23}} & \textbf{\textsc{MATH500}} & \textbf{\textsc{GPQA}} & \textbf{\textsc{Minerva}} & \textbf{\textsc{Avg.}} \\
\midrule
1000 & 33.33&	16.67&	72.50&	84.00&	31.99&	40.91&	46.57 \\
\midrule
2000 & 33.33&	30.00&	77.50&	86.80&	27.21&	41.92&	49.46 \\
\midrule
3000 & 30.00&	30.00&	72.50&	84.60&	29.04&	33.84&	46.66 \\
\midrule
4000 & 33.33&	23.33&	65.00&	84.20&	29.04&	34.34&	44.88 \\
\midrule
5000 & 33.33&	23.33&	67.50&	84.40&	28.68&	32.32&	44.93 \\
\midrule
6000 & 20.00&	20.00&	72.50&	84.20&	28.68&	31.31&	42.78 \\
\midrule
7000 & 23.33&	20.00&	67.50&	81.40&	30.88&	37.88&	43.50 \\
\midrule
8000 & 33.33&	23.33&	72.50&	80.40&	26.10&	35.86&	45.25 \\
\midrule
9000 & 30.00&	23.33&	70.00&	83.60&	27.94&	30.81&	44.28 \\
\bottomrule
\end{tabular}
}
\caption{\textbf{Performance of Resa-STILL2OpenR1}\; Each epoch contains 4911 Steps.}
\end{table}

\begin{table}[h]
\centering
\resizebox{.9\textwidth}{!}{
\begin{tabular}{c|cccccccc|c}
\toprule
\rowcolor{LightGreen}
\textbf{\textsc{Checkpoint Steps}} & \textbf{\textsc{AIME24}} & \textbf{\textsc{AIME25}} & \textbf{\textsc{AMC23}} & \textbf{\textsc{MATH500}} & \textbf{\textsc{GPQA}} & \textbf{\textsc{Minerva}} & \textbf{\textsc{Avg.}} \\
\midrule
1000 & 36.67&	20.00&	82.50&	83.40&	31.25&	33.33&	47.86 \\
\midrule
1500 & 36.67&	16.67&	67.50&	86.20&	31.25&	34.85&	45.52 \\
\midrule
2000 & 40.00&	20.00&	72.50&	84.60&	30.15&	32.32&	46.60 \\
\midrule
2500 & 26.67&	16.67&	72.50&	84.60&	26.84&	35.86&	43.86 \\
\bottomrule
\end{tabular}
}
\caption{\textbf{Performance of Resa-STILL-Qwen-Math-Adapter}\; Each epoch contains 1448 Steps.}
\end{table}

\begin{table}[h]
\centering
\resizebox{.9\textwidth}{!}{
\begin{tabular}{c|cccccccc|c}
\toprule
\rowcolor{LightGreen}
\textbf{\textsc{Checkpoint Steps}} & \textbf{\textsc{AIME24}} & \textbf{\textsc{AIME25}} & \textbf{\textsc{AMC23}} & \textbf{\textsc{MATH500}} & \textbf{\textsc{GPQA}} & \textbf{\textsc{Minerva}} & \textbf{\textsc{Avg.}} \\
\midrule
1000 & 30.00&	30.00&	72.50&	85.60&	31.25&	35.86&	47.54 \\
\midrule
1500 & 20.00&	20.00&	72.50&	83.00&	30.15&	35.35&	43.50 \\
\midrule
2000 & 26.67&	30.00&	67.50&	84.60&	25.74&	32.83&	44.56 \\
\midrule
2500 & 30.00&	16.67&	70.00&	83.60&	29.78&	34.34&	44.07 \\
\bottomrule
\end{tabular}
}
\caption{\textbf{Performance of Resa-STILL-Qwen-Adapter}\; Each epoch contains 1448 Steps.}
\end{table}

\clearpage
\subsection{Full Results of Table~\ref{tab:layer ablation}}
\label{app:tab layer ablation}

In the following tables, we present the full performance evaluation results of models in Table~\ref{tab:layer ablation}.

\begin{table}[h]
\centering
\resizebox{.9\textwidth}{!}{
\begin{tabular}{c|cccccccc|c}
\toprule
\rowcolor{LightGreen}
\textbf{\textsc{Checkpoint Steps}} & \textbf{\textsc{AIME24}} & \textbf{\textsc{AIME25}} & \textbf{\textsc{AMC23}} & \textbf{\textsc{MATH500}} & \textbf{\textsc{GPQA}} & \textbf{\textsc{Minerva}} & \textbf{\textsc{Avg.}} \\
\midrule
1000 & 26.67&	20.00&	70.00&	85.00&	29.41&	32.32&	43.90 \\
\midrule
1500 & 26.67&	16.67&	70.00&	82.40&	26.84&	31.31&	42.31 \\
\midrule
2000 & 26.67&	30.00&	80.00&	83.20&	29.78&	37.37&	47.84 \\
\midrule
2500 & 16.67&	23.33&	67.50&	86.00&	32.35&	34.34&	43.37 \\
\bottomrule
\end{tabular}
}
\caption{\textbf{Performance of Resa-STILL-2nd-Layer}\; Each epoch contains 1448 Steps.}
\end{table}

\begin{table}[h]
\centering
\resizebox{.9\textwidth}{!}{
\begin{tabular}{c|cccccccc|c}
\toprule
\rowcolor{LightGreen}
\textbf{\textsc{Checkpoint Steps}} & \textbf{\textsc{AIME24}} & \textbf{\textsc{AIME25}} & \textbf{\textsc{AMC23}} & \textbf{\textsc{MATH500}} & \textbf{\textsc{GPQA}} & \textbf{\textsc{Minerva}} & \textbf{\textsc{Avg.}} \\
\midrule
1000 & 20.00&	20.00&	82.50&	84.60&	24.26&	32.32&	43.95 \\
\midrule
1500 & 26.67&	36.67&	70.00&	83.40&	28.31&	33.84&	46.48 \\
\midrule
2000 & 33.33&	23.33&	70.00&	83.40&	28.68&	33.84&	45.43 \\
\midrule
2500 & 26.67&	16.67&	72.50&	84.60&	31.25&	30.30&	43.66 \\
\bottomrule
\end{tabular}
}
\caption{\textbf{Performance of Resa-STILL-3rd-Layer}\; Each epoch contains 1448 Steps.}
\end{table}

\begin{table}[h]
\centering
\resizebox{.9\textwidth}{!}{
\begin{tabular}{c|cccccccc|c}
\toprule
\rowcolor{LightGreen}
\textbf{\textsc{Checkpoint Steps}} & \textbf{\textsc{AIME24}} & \textbf{\textsc{AIME25}} & \textbf{\textsc{AMC23}} & \textbf{\textsc{MATH500}} & \textbf{\textsc{GPQA}} & \textbf{\textsc{Minerva}} & \textbf{\textsc{Avg.}} \\
\midrule
1000 & 26.67&	16.67&	67.50&	84.20&	28.31&	40.91&	44.04 \\
\midrule
1500 & 33.33&	20.00&	80.00&	83.80&	28.68&	36.87&	47.11 \\
\midrule
2000 & 36.67&	23.33&	70.00&	83.20&	27.57&	34.85&	45.94 \\
\midrule
2500 & 40.00&	20.00&	72.50&	83.80&	30.88&	31.31&	46.42 \\
\bottomrule
\end{tabular}
}
\caption{\textbf{Performance of Resa-STILL-4th-Layer}\; Each epoch contains 1448 Steps.}
\end{table}

\begin{table}[h]
\centering
\resizebox{.9\textwidth}{!}{
\begin{tabular}{c|cccccccc|c}
\toprule
\rowcolor{LightGreen}
\textbf{\textsc{Checkpoint Steps}} & \textbf{\textsc{AIME24}} & \textbf{\textsc{AIME25}} & \textbf{\textsc{AMC23}} & \textbf{\textsc{MATH500}} & \textbf{\textsc{GPQA}} & \textbf{\textsc{Minerva}} & \textbf{\textsc{Avg.}} \\
\midrule
1000 & 26.67&	20.00&	75.00&	80.80&	29.41&	32.32&	44.03 \\
\midrule
1500 & 30.00&	16.67&	72.50&	82.80&	30.51&	30.81&	43.88 \\
\midrule
2000 & 40.00&	23.33&	70.00&	83.20&	26.84&	44.95&	48.05 \\
\midrule
2500 & 33.33&	26.67&	70.00&	83.60&	29.41&	31.31&	45.72 \\
\bottomrule
\end{tabular}
}
\caption{\textbf{Performance of Resa-STILL-5th-Layer}\; Each epoch contains 1448 Steps.}
\end{table}

\begin{table}[h]
\centering
\resizebox{.9\textwidth}{!}{
\begin{tabular}{c|cccccccc|c}
\toprule
\rowcolor{LightGreen}
\textbf{\textsc{Checkpoint Steps}} & \textbf{\textsc{AIME24}} & \textbf{\textsc{AIME25}} & \textbf{\textsc{AMC23}} & \textbf{\textsc{MATH500}} & \textbf{\textsc{GPQA}} & \textbf{\textsc{Minerva}} & \textbf{\textsc{Avg.}} \\
\midrule
1000 & 20.00&	20.00&	75.00&	85.60&	27.57&	39.90&	44.68 \\
\midrule
1500 & 30.00&	20.00&	80.00&	83.20&	30.88&	31.31&	45.90 \\
\midrule
2000 & 33.33&	23.33&	72.50&	83.40&	31.62&	35.35&	46.59 \\
\midrule
2500 & 30.00&	16.67&	72.50&	82.60&	27.94&	38.38&	44.68 \\
\bottomrule
\end{tabular}
}
\caption{\textbf{Performance of Resa-STILL-6th-Layer}\; Each epoch contains 1448 Steps.}
\end{table}

\begin{table}[h]
\centering
\resizebox{.9\textwidth}{!}{
\begin{tabular}{c|cccccccc|c}
\toprule
\rowcolor{LightGreen}
\textbf{\textsc{Checkpoint Steps}} & \textbf{\textsc{AIME24}} & \textbf{\textsc{AIME25}} & \textbf{\textsc{AMC23}} & \textbf{\textsc{MATH500}} & \textbf{\textsc{GPQA}} & \textbf{\textsc{Minerva}} & \textbf{\textsc{Avg.}} \\
\midrule
1000 & 33.33&	20.00&	72.50&	83.40&	31.62&	30.30&	45.19 \\
\midrule
1500 & 26.67&	26.67&	77.50&	81.60&	27.94&	35.86&	46.04 \\
\midrule
2000 & 23.33&	20.00&	70.00&	82.60&	29.78&	40.91&	44.44 \\
\midrule
2500 & 30.00&	33.33&	67.50&	83.20&	24.63&	36.36&	45.84 \\
\bottomrule
\end{tabular}
}
\caption{\textbf{Performance of Resa-STILL-7th-Layer}\; Each epoch contains 1448 Steps.}
\end{table}

\begin{table}[h]
\centering
\resizebox{.9\textwidth}{!}{
\begin{tabular}{c|cccccccc|c}
\toprule
\rowcolor{LightGreen}
\textbf{\textsc{Checkpoint Steps}} & \textbf{\textsc{AIME24}} & \textbf{\textsc{AIME25}} & \textbf{\textsc{AMC23}} & \textbf{\textsc{MATH500}} & \textbf{\textsc{GPQA}} & \textbf{\textsc{Minerva}} & \textbf{\textsc{Avg.}} \\
\midrule
1000 & 20.00&	23.33&	82.50&	85.20&	29.78&	33.33&	45.69 \\
\midrule
1500 & 16.67&	23.33&	70.00&	82.60&	27.21&	34.34&	42.36 \\
\midrule
2000 & 30.00&	20.00&	70.00&	83.20&	28.31&	37.88&	44.90 \\
\midrule
2500 & 16.67&	20.00&	72.50&	81.00&	29.04&	28.79&	41.33 \\
\bottomrule
\end{tabular}
}
\caption{\textbf{Performance of Resa-STILL-8th-Layer}\; Each epoch contains 1448 Steps.}
\end{table}

\begin{table}[h]
\centering
\resizebox{.9\textwidth}{!}{
\begin{tabular}{c|cccccccc|c}
\toprule
\rowcolor{LightGreen}
\textbf{\textsc{Checkpoint Steps}} & \textbf{\textsc{AIME24}} & \textbf{\textsc{AIME25}} & \textbf{\textsc{AMC23}} & \textbf{\textsc{MATH500}} & \textbf{\textsc{GPQA}} & \textbf{\textsc{Minerva}} & \textbf{\textsc{Avg.}} \\
\midrule
1000 & 20.00&	36.67&	77.50&	84.20&	25.37&	37.88&	46.94 \\
\midrule
1500 & 36.67&	20.00&	75.00&	83.80&	28.31&	34.34&	46.35 \\
\midrule
2000 & 36.67&	20.00&	80.00&	84.20&	27.57&	34.34&	47.13 \\
\midrule
2500 & 26.67&	26.67&	67.50&	83.80&	27.57&	38.89&	45.18 \\
\bottomrule
\end{tabular}
}
\caption{\textbf{Performance of Resa-STILL-9th-Layer}\; Each epoch contains 1448 Steps.}
\end{table}

\begin{table}[h]
\centering
\resizebox{.9\textwidth}{!}{
\begin{tabular}{c|cccccccc|c}
\toprule
\rowcolor{LightGreen}
\textbf{\textsc{Checkpoint Steps}} & \textbf{\textsc{AIME24}} & \textbf{\textsc{AIME25}} & \textbf{\textsc{AMC23}} & \textbf{\textsc{MATH500}} & \textbf{\textsc{GPQA}} & \textbf{\textsc{Minerva}} & \textbf{\textsc{Avg.}} \\
\midrule
1000 & 26.67&	23.33&	75.00&	84.60&	27.21&	33.84&	45.11 \\
\midrule
1500 & 36.67&	23.33&	67.50&	84.80&	29.78&	37.37&	46.58 \\
\midrule
2000 & 23.33&	23.33&	75.00&	84.20&	29.41&	25.25&	43.42 \\
\midrule
2500 & 36.67&	23.33&	70.00&	83.20&	29.41&	36.36&	46.50 \\
\bottomrule
\end{tabular}
}
\caption{\textbf{Performance of Resa-STILL-10th-Layer}\; Each epoch contains 1448 Steps.}
\end{table}

\begin{table}[h]
\centering
\resizebox{.9\textwidth}{!}{
\begin{tabular}{c|cccccccc|c}
\toprule
\rowcolor{LightGreen}
\textbf{\textsc{Checkpoint Steps}} & \textbf{\textsc{AIME24}} & \textbf{\textsc{AIME25}} & \textbf{\textsc{AMC23}} & \textbf{\textsc{MATH500}} & \textbf{\textsc{GPQA}} & \textbf{\textsc{Minerva}} & \textbf{\textsc{Avg.}} \\
\midrule
1000 & 26.67&	36.67&	77.50&	83.80&	31.25&	32.83&	48.12 \\
\midrule
1500 & 26.67&	33.33&	75.00&	84.60&	30.51&	34.34&	47.41 \\
\midrule
2000 & 30.00&	23.33&	67.50&	84.60&	31.99&	45.45&	47.15 \\
\midrule
2500 & 20.00&	26.67&	72.50&	83.60&	33.09&	33.84&	44.95 \\
\bottomrule
\end{tabular}
}
\caption{\textbf{Performance of Resa-STILL-11th-Layer}\; Each epoch contains 1448 Steps.}
\end{table}

\begin{table}[h]
\centering
\resizebox{.9\textwidth}{!}{
\begin{tabular}{c|cccccccc|c}
\toprule
\rowcolor{LightGreen}
\textbf{\textsc{Checkpoint Steps}} & \textbf{\textsc{AIME24}} & \textbf{\textsc{AIME25}} & \textbf{\textsc{AMC23}} & \textbf{\textsc{MATH500}} & \textbf{\textsc{GPQA}} & \textbf{\textsc{Minerva}} & \textbf{\textsc{Avg.}} \\
\midrule
1000 & 36.67&	23.33&	70.00&	82.20&	29.04&	34.34&	45.93 \\
\midrule
1500 & 20.00&	23.33&	72.50&	84.60&	26.84&	40.40&	44.61 \\
\midrule
2000 & 30.00&	30.00&	72.50&	82.60&	30.51&	32.83&	46.41 \\
\midrule
2500 & 30.00&	26.67&	77.50&	84.60&	31.99&	34.85&	47.60 \\
\bottomrule
\end{tabular}
}
\caption{\textbf{Performance of Resa-STILL-12th-Layer}\; Each epoch contains 1448 Steps.}
\end{table}

\begin{table}[h]
\centering
\resizebox{.9\textwidth}{!}{
\begin{tabular}{c|cccccccc|c}
\toprule
\rowcolor{LightGreen}
\textbf{\textsc{Checkpoint Steps}} & \textbf{\textsc{AIME24}} & \textbf{\textsc{AIME25}} & \textbf{\textsc{AMC23}} & \textbf{\textsc{MATH500}} & \textbf{\textsc{GPQA}} & \textbf{\textsc{Minerva}} & \textbf{\textsc{Avg.}} \\
\midrule
1000 & 20.00&	33.33&	75.00&	82.60&	29.78&	33.33&	45.67 \\
\midrule
1500 & 33.33&	23.33&	75.00&	82.80&	30.88&	30.81&	46.03 \\
\midrule
2000 & 33.33&	33.33&	75.00&	83.80&	29.41&	28.79&	47.28 \\
\midrule
2500 & 30.00&	23.33&	77.50&	84.20&	26.47&	33.33&	45.81 \\
\bottomrule
\end{tabular}
}
\caption{\textbf{Performance of Resa-STILL-13th-Layer}\; Each epoch contains 1448 Steps.}
\end{table}

\begin{table}[h]
\centering
\resizebox{.9\textwidth}{!}{
\begin{tabular}{c|cccccccc|c}
\toprule
\rowcolor{LightGreen}
\textbf{\textsc{Checkpoint Steps}} & \textbf{\textsc{AIME24}} & \textbf{\textsc{AIME25}} & \textbf{\textsc{AMC23}} & \textbf{\textsc{MATH500}} & \textbf{\textsc{GPQA}} & \textbf{\textsc{Minerva}} & \textbf{\textsc{Avg.}} \\
\midrule
1000 & 33.33&	20.00&	77.50&	83.60&	28.68&	29.80&	45.48 \\
\midrule
1500 & 26.67&	26.67&	70.00&	84.80&	26.84&	30.81&	44.30 \\
\midrule
2000 & 23.33&	16.67&	72.50&	83.80&	28.31&	34.34&	43.16 \\
\midrule
2500 & 33.33&	23.33&	65.00&	82.80&	30.88&	32.32&	44.61 \\
\bottomrule
\end{tabular}
}
\caption{\textbf{Performance of Resa-STILL-14th-Layer}\; Each epoch contains 1448 Steps.}
\end{table}

\begin{table}[h]
\centering
\resizebox{.9\textwidth}{!}{
\begin{tabular}{c|cccccccc|c}
\toprule
\rowcolor{LightGreen}
\textbf{\textsc{Checkpoint Steps}} & \textbf{\textsc{AIME24}} & \textbf{\textsc{AIME25}} & \textbf{\textsc{AMC23}} & \textbf{\textsc{MATH500}} & \textbf{\textsc{GPQA}} & \textbf{\textsc{Minerva}} & \textbf{\textsc{Avg.}} \\
\midrule
1000 & 36.67&	23.33&	80.00&	84.80&	30.15&	38.89&	48.97 \\
\midrule
1500 & 26.67&	26.67&	72.50&	84.40&	30.51&	31.82&	45.43 \\
\midrule
2000 & 36.67&	20.00&	72.50&	82.00&	29.04&	36.36&	46.10 \\
\midrule
2500 & 30.00&	16.67&	67.50&	83.80&	32.72&	36.87&	44.59 \\
\bottomrule
\end{tabular}
}
\caption{\textbf{Performance of Resa-STILL-15th-Layer}\; Each epoch contains 1448 Steps.}
\end{table}

\begin{table}[h]
\centering
\resizebox{.9\textwidth}{!}{
\begin{tabular}{c|cccccccc|c}
\toprule
\rowcolor{LightGreen}
\textbf{\textsc{Checkpoint Steps}} & \textbf{\textsc{AIME24}} & \textbf{\textsc{AIME25}} & \textbf{\textsc{AMC23}} & \textbf{\textsc{MATH500}} & \textbf{\textsc{GPQA}} & \textbf{\textsc{Minerva}} & \textbf{\textsc{Avg.}} \\
\midrule
1000 & 23.33&	16.67&	72.50&	85.40&	30.15&	38.89&	44.49 \\
\midrule
1500 & 30.00&	20.00&	65.00&	83.40&	29.78&	32.83&	43.50 \\
\midrule
2000 & 30.00&	20.00&	80.00&	85.00&	30.15&	34.85&	46.67 \\
\midrule
2500 & 30.00&	20.00&	70.00&	82.80&	30.15&	38.38&	45.22 \\
\bottomrule
\end{tabular}
}
\caption{\textbf{Performance of Resa-STILL-16th-Layer}\; Each epoch contains 1448 Steps.}
\end{table}

\begin{table}[h]
\centering
\resizebox{.9\textwidth}{!}{
\begin{tabular}{c|cccccccc|c}
\toprule
\rowcolor{LightGreen}
\textbf{\textsc{Checkpoint Steps}} & \textbf{\textsc{AIME24}} & \textbf{\textsc{AIME25}} & \textbf{\textsc{AMC23}} & \textbf{\textsc{MATH500}} & \textbf{\textsc{GPQA}} & \textbf{\textsc{Minerva}} & \textbf{\textsc{Avg.}} \\
\midrule
1000 & 33.33&	30.00&	67.50&	85.40&	30.88&	36.36&	47.25 \\
\midrule
1500 & 33.33&	13.33&	67.50&	83.00&	30.88&	35.86&	43.98 \\
\midrule
2000 & 36.67&	23.33&	65.00&	83.20&	31.25&	34.85&	45.72 \\
\midrule
2500 & 43.33&	16.67&	77.50&	83.80&	30.88&	36.36&	48.09 \\
\bottomrule
\end{tabular}
}
\caption{\textbf{Performance of Resa-STILL-17th-Layer}\; Each epoch contains 1448 Steps.}
\end{table}

\begin{table}[h]
\centering
\resizebox{.9\textwidth}{!}{
\begin{tabular}{c|cccccccc|c}
\toprule
\rowcolor{LightGreen}
\textbf{\textsc{Checkpoint Steps}} & \textbf{\textsc{AIME24}} & \textbf{\textsc{AIME25}} & \textbf{\textsc{AMC23}} & \textbf{\textsc{MATH500}} & \textbf{\textsc{GPQA}} & \textbf{\textsc{Minerva}} & \textbf{\textsc{Avg.}} \\
\midrule
1000 & 43.33&	20.00&	77.50&	84.00&	33.82&	37.88&	49.42 \\
\midrule
1500 & 26.67&	20.00&	75.00&	83.00&	29.41&	35.86&	44.99 \\
\midrule
2000 & 23.33&	23.33&	75.00&	83.40&	33.82&	38.38&	46.21 \\
\midrule
2500 & 43.33&	23.33&	72.50&	84.40&	29.04&	31.82&	47.40 \\
\bottomrule
\end{tabular}
}
\caption{\textbf{Performance of Resa-STILL-18th-Layer}\; Each epoch contains 1448 Steps.}
\end{table}

\begin{table}[h]
\centering
\resizebox{.9\textwidth}{!}{
\begin{tabular}{c|cccccccc|c}
\toprule
\rowcolor{LightGreen}
\textbf{\textsc{Checkpoint Steps}} & \textbf{\textsc{AIME24}} & \textbf{\textsc{AIME25}} & \textbf{\textsc{AMC23}} & \textbf{\textsc{MATH500}} & \textbf{\textsc{GPQA}} & \textbf{\textsc{Minerva}} & \textbf{\textsc{Avg.}} \\
\midrule
1000 & 40.00&	13.33&	75.00&	84.20&	27.94&	38.38&	46.48 \\
\midrule
1500 & 30.00&	23.33&	72.50&	83.20&	30.15&	35.35&	45.76 \\
\midrule
2000 & 33.33&	30.00&	72.50&	84.20&	29.04&	36.87&	47.66 \\
\midrule
2500 & 26.67&	23.33&	75.00&	84.80&	28.68&	35.35&	45.64 \\
\bottomrule
\end{tabular}
}
\caption{\textbf{Performance of Resa-STILL-19th-Layer}\; Each epoch contains 1448 Steps.}
\end{table}

\begin{table}[h]
\centering
\resizebox{.9\textwidth}{!}{
\begin{tabular}{c|cccccccc|c}
\toprule
\rowcolor{LightGreen}
\textbf{\textsc{Checkpoint Steps}} & \textbf{\textsc{AIME24}} & \textbf{\textsc{AIME25}} & \textbf{\textsc{AMC23}} & \textbf{\textsc{MATH500}} & \textbf{\textsc{GPQA}} & \textbf{\textsc{Minerva}} & \textbf{\textsc{Avg.}} \\
\midrule
1000 & 26.67&	33.33&	70.00&	83.60&	27.21&	31.82&	45.44 \\
\midrule
1500 & 33.33&	23.33&	75.00&	85.00&	29.04&	29.29&	45.83 \\
\midrule
2000 & 20.00&	30.00&	65.00&	83.60&	28.31&	27.27&	42.36 \\
\midrule
2500 & 40.00&	20.00&	67.50&	82.80&	27.94&	33.84&	45.35 \\
\bottomrule
\end{tabular}
}
\caption{\textbf{Performance of Resa-STILL-20th-Layer}\; Each epoch contains 1448 Steps.}
\end{table}

\begin{table}[h]
\centering
\resizebox{.9\textwidth}{!}{
\begin{tabular}{c|cccccccc|c}
\toprule
\rowcolor{LightGreen}
\textbf{\textsc{Checkpoint Steps}} & \textbf{\textsc{AIME24}} & \textbf{\textsc{AIME25}} & \textbf{\textsc{AMC23}} & \textbf{\textsc{MATH500}} & \textbf{\textsc{GPQA}} & \textbf{\textsc{Minerva}} & \textbf{\textsc{Avg.}} \\
\midrule
1000 & 30.00&	33.33&	75.00&	84.60&	27.57&	36.87&	47.90 \\
\midrule
1500 & 33.33&	16.67&	72.50&	84.20&	30.51&	30.81&	44.67 \\
\midrule
2000 & 40.00&	20.00&	62.50&	81.80&	29.41&	34.85&	44.76 \\
\midrule
2500 & 23.33&	30.00&	70.00&	84.00&	30.88&	33.84&	45.34 \\
\bottomrule
\end{tabular}
}
\caption{\textbf{Performance of Resa-STILL-21st-Layer}\; Each epoch contains 1448 Steps.}
\end{table}

\begin{table}[h]
\centering
\resizebox{.9\textwidth}{!}{
\begin{tabular}{c|cccccccc|c}
\toprule
\rowcolor{LightGreen}
\textbf{\textsc{Checkpoint Steps}} & \textbf{\textsc{AIME24}} & \textbf{\textsc{AIME25}} & \textbf{\textsc{AMC23}} & \textbf{\textsc{MATH500}} & \textbf{\textsc{GPQA}} & \textbf{\textsc{Minerva}} & \textbf{\textsc{Avg.}} \\
\midrule
1000 & 33.33&	23.33&	75.00&	82.20&	31.99&	34.34&	46.70 \\
\midrule
1500 & 30.00&	26.67&	70.00&	83.20&	31.62&	35.86&	46.23 \\
\midrule
2000 & 23.33&	16.67&	70.00&	83.80&	31.25&	34.85&	43.32 \\
\midrule
2500 & 23.33&	20.00&	72.50&	85.80&	29.04&	39.39&	45.01 \\
\bottomrule
\end{tabular}
}
\caption{\textbf{Performance of Resa-STILL-22nd-Layer}\; Each epoch contains 1448 Steps.}
\end{table}

\begin{table}[h]
\centering
\resizebox{.9\textwidth}{!}{
\begin{tabular}{c|cccccccc|c}
\toprule
\rowcolor{LightGreen}
\textbf{\textsc{Checkpoint Steps}} & \textbf{\textsc{AIME24}} & \textbf{\textsc{AIME25}} & \textbf{\textsc{AMC23}} & \textbf{\textsc{MATH500}} & \textbf{\textsc{GPQA}} & \textbf{\textsc{Minerva}} & \textbf{\textsc{Avg.}} \\
\midrule
1000 & 30.00&	20.00&	72.50&	83.80&	28.31&	32.32&	44.49 \\
\midrule
1500 & 30.00&	20.00&	80.00&	82.00&	29.04&	35.35&	46.07 \\
\midrule
2000 & 23.33&	23.33&	60.00&	84.80&	27.57&	37.37&	42.73 \\
\midrule
2500 & 23.33&	23.33&	67.50&	85.00&	29.78&	43.43&	45.40 \\
\bottomrule
\end{tabular}
}
\caption{\textbf{Performance of Resa-STILL-23rd-Layer}\; Each epoch contains 1448 Steps.}
\end{table}

\begin{table}[h]
\centering
\resizebox{.9\textwidth}{!}{
\begin{tabular}{c|cccccccc|c}
\toprule
\rowcolor{LightGreen}
\textbf{\textsc{Checkpoint Steps}} & \textbf{\textsc{AIME24}} & \textbf{\textsc{AIME25}} & \textbf{\textsc{AMC23}} & \textbf{\textsc{MATH500}} & \textbf{\textsc{GPQA}} & \textbf{\textsc{Minerva}} & \textbf{\textsc{Avg.}} \\
\midrule
1000 & 36.67&	20.00&	77.50&	83.80&	30.51&	29.80&	46.38 \\
\midrule
1500 & 20.00&	16.67&	65.00&	85.20&	29.41&	34.85&	41.86 \\
\midrule
2000 & 26.67&	33.33&	67.50&	85.40&	30.15&	33.33&	46.06 \\
\midrule
2500 & 33.33&	16.67&	75.00&	82.80&	29.78&	33.33&	45.15 \\
\bottomrule
\end{tabular}
}
\caption{\textbf{Performance of Resa-STILL-24th-Layer}\; Each epoch contains 1448 Steps.}
\end{table}

\begin{table}[h]
\centering
\resizebox{.9\textwidth}{!}{
\begin{tabular}{c|cccccccc|c}
\toprule
\rowcolor{LightGreen}
\textbf{\textsc{Checkpoint Steps}} & \textbf{\textsc{AIME24}} & \textbf{\textsc{AIME25}} & \textbf{\textsc{AMC23}} & \textbf{\textsc{MATH500}} & \textbf{\textsc{GPQA}} & \textbf{\textsc{Minerva}} & \textbf{\textsc{Avg.}} \\
\midrule
1000 & 40.00&	30.00&	70.00&	84.20&	28.68&	37.88&	48.46 \\
\midrule
1500 & 30.00&	26.67&	67.50&	83.20&	29.41&	35.86&	45.44 \\
\midrule
2000 & 23.33&	16.67&	75.00&	82.80&	30.88&	36.36&	44.17 \\
\midrule
2500 & 30.00&	26.67&	67.50&	84.60&	27.57&	34.34&	45.11 \\
\bottomrule
\end{tabular}
}
\caption{\textbf{Performance of Resa-STILL-25th-Layer}\; Each epoch contains 1448 Steps.}
\end{table}

\begin{table}[h]
\centering
\resizebox{.9\textwidth}{!}{
\begin{tabular}{c|cccccccc|c}
\toprule
\rowcolor{LightGreen}
\textbf{\textsc{Checkpoint Steps}} & \textbf{\textsc{AIME24}} & \textbf{\textsc{AIME25}} & \textbf{\textsc{AMC23}} & \textbf{\textsc{MATH500}} & \textbf{\textsc{GPQA}} & \textbf{\textsc{Minerva}} & \textbf{\textsc{Avg.}} \\
\midrule
1000 & 36.67&	20.00&	65.00&	85.60&	30.88&	36.87&	45.84 \\
\midrule
1500 & 26.67&	16.67&	75.00&	83.60&	30.15&	33.33&	44.24 \\
\midrule
2000 & 26.67&	23.33&	67.50&	84.40&	27.21&	35.86&	44.16 \\
\midrule
2500 & 30.00&	23.33&	70.00&	83.00&	31.62&	36.87&	45.80 \\
\bottomrule
\end{tabular}
}
\caption{\textbf{Performance of Resa-STILL-26th-Layer}\; Each epoch contains 1448 Steps.}
\end{table}

\begin{table}[h]
\centering
\resizebox{.9\textwidth}{!}{
\begin{tabular}{c|cccccccc|c}
\toprule
\rowcolor{LightGreen}
\textbf{\textsc{Checkpoint Steps}} & \textbf{\textsc{AIME24}} & \textbf{\textsc{AIME25}} & \textbf{\textsc{AMC23}} & \textbf{\textsc{MATH500}} & \textbf{\textsc{GPQA}} & \textbf{\textsc{Minerva}} & \textbf{\textsc{Avg.}} \\
\midrule
1000 & 16.67&	26.67&	65.00&	82.80&	30.51&	34.85&	42.75 \\
\midrule
1500 & 33.33&	23.33&	72.50&	82.80&	31.62&	36.87&	46.74 \\
\midrule
2000 & 36.67&	30.00&	62.50&	83.00&	27.94&	33.84&	45.66 \\
\midrule
2500 & 30.00&	36.67&	67.50&	83.00&	27.94&	40.40&	47.59 \\
\bottomrule
\end{tabular}
}
\caption{\textbf{Performance of Resa-STILL-27th-Layer}\; Each epoch contains 1448 Steps.}
\end{table}

\appendix
\end{document}